\PassOptionsToPackage{round}{natbib}
\documentclass{article}
\usepackage[preprint]{neurips_2024}

\usepackage{fontspec}
\newfontfamily\FBserif[Path=./]{NotoSerif-sub.ttf}
\newfontfamily\FBheb[Path=./]{NotoSansHebrew-sub.ttf}
\newfontfamily\FBarab[Path=./,Script=Arabic]{NotoNaskhArabic-sub.ttf}
\newfontfamily\FBdev[Path=./,Script=Devanagari]{NotoSansDevanagari-sub.ttf}
\newfontfamily\FBori[Path=./,Script=Oriya]{NotoSansOriya-sub.ttf}
\newfontfamily\FBtib[Path=./,Script=Tibetan]{NotoSerifTibetan-sub.ttf}
\newfontfamily\FBcjk[Path=./]{NotoSansCJKjp-Regular-subset.ttf}
\newfontfamily\FBsym[Path=./]{NotoSansSymbols2-sub.ttf}
\newfontfamily\FBmathf[Path=./]{NotoSansMath-sub.ttf}
\newcommand{\emojipng}[1]{\raisebox{-0.25ex}{\includegraphics[height=1.7ex]{#1.png}}}

\usepackage{booktabs}
\usepackage{tabularx}
\newcolumntype{Y}{>{\raggedright\arraybackslash}X}
\usepackage{longtable}
\usepackage{array}
\usepackage{graphicx}
\usepackage{amsmath}
\usepackage{amssymb}
\usepackage{xfrac}
\usepackage{enumitem}
\usepackage{multicol}
\providecommand{\tightlist}{\setlength{\itemsep}{0pt}\setlength{\parskip}{0pt}}
\usepackage{calc}


\usepackage{placeins}
\usepackage{caption}
\makeatletter
\renewcommand{\@noticestring}{Preprint. Licensed under CC BY 4.0}
\makeatother

\usepackage[colorlinks=true,allcolors=blue,breaklinks=true]{hyperref}
\hypersetup{
  pdftitle={Replicating the Geometry of Emotion Representations in a Base Open-Weights Model},
  pdfauthor={Adam Hollowell},
  pdfsubject={Mechanistic interpretability; emotion representations; replication},
  pdfkeywords={interpretability, emotion vectors, Gemma 2, replication, affective circumplex}}
\usepackage{url}

\title{Replicating the Geometry of\\Emotion Representations\\in a Base Open-Weights Model}
\author{%
  Adam Hollowell\\
  University of North Carolina at Chapel Hill\\
  \texttt{adam.hollowell@unc.edu}
}

\begin{document}
\maketitle

\begin{abstract}
Sofroniew et al.~(2026) report that emotion concepts in Claude Sonnet 4.5 are represented as vectors whose geometry mirrors human affect psychology. We replicate the representational core of that study on the base pretrained model \texttt{google/gemma-2-27b}, inheriting every disclosed parameter, resolving unspecified steps by disclosed rules, and changing only the subject model. From 205,200 newly generated Claude Sonnet 4.5 stories matching the original corpus design, we extract 171 emotion vectors and recover the core results. The leading principal components form an affective circumplex (PC1 carries 26.7\% of variance against the original study's \textasciitilde{}27\%, PC2 13.4\% against \textasciitilde{}14\%), emotions cluster into similar intuitive families, and the geometry holds across a broad late-middle band. The valence axis aligns with human norms (r = 0.72, against 0.81) and is stable across scales and depth. Arousal aligns at r = 0.67 (against 0.66) but only at the full 171-emotion scale and late depth, so we do not classify it as replicated. Extending the original analysis, a 46-layer sweep locates a sharp seam at L22–26, where the geometry consolidates and the vocabulary readout becomes legible. An embedding-layer baseline finds much of the geometry already present in the static token embeddings, with arousal as the exception. On top-activating held-out text, the geometry predicts token-level co-activation at r = 0.907. At least 52\% of vectors peak on structurally non-conceptual tokens, a measured floor for max-activation confounds. Even when a document contains the vector's emotion word, the peak lands on that word only 6.1\% of the time. Because the subject is a base model and the stimuli are Claude-generated fiction, the recovered structure is a property of the pretrained representation of Claude-rendered emotion. The original’s causal and assistant-facing analyses are out of scope. Code and data are released.

\end{abstract}

\section{Introduction}\label{introduction}

Sofroniew et al.~(\citeyear{sofroniew2026emotion}; arXiv:2604.07729)
show, in Claude Sonnet 4.5, that emotion vectors extracted from the
model's activations over 205,200 short stories organize into a geometry
that mirrors human affect psychology. Valence and arousal define the
leading axes, emotions cluster into intuitive families, and the geometry
holds across middle-to-late layers. \citet{sofroniew2026emotion} expect
that these broad findings should generalize beyond one post-trained
frontier model, while the details will likely vary across families,
sizes, and training procedures. A direct test of that expectation would
be a replication study that inherits their extraction and
characterization pipeline and changes only the model.

This paper reports such a replication on the base pretrained model
\texttt{google/gemma-2-27b}. For every decision
\citet{sofroniew2026emotion} disclose, we inherit their choice: the list
of 171 emotions, the verbatim story elicitation prompt, the $\sfrac{2}{3}$-depth
analysis layer, the token-50 pooling rule, the difference-of-means
construction, the 50\%-variance confound-cleanup threshold, the
300-story validation design, the Russell--Mehrabian benchmark and its
45-emotion overlap, k = 10 for clustering, and the held-out corpus
sources (Table~\ref*{tab:1}). Where the original study describes a step without
fully specifying it---such as the sign convention for principal
components or the coding procedure behind its validation---we resolve it
by a disclosed rule, marked in Table~\ref*{tab:1}. Running the pipeline on open
weights also requires choices with no counterpart in the original study,
and we document each of those choices in \S3. We label analyses with no
counterpart as extensions where they occur.

We did not tune the choices inherited from \citet{sofroniew2026emotion}
to make Gemma agree with Claude. That discipline lets agreement count as
convergent evidence and divergence as a cross-model difference rather
than a pipeline artifact. (Relatedly, \citealt{jeong2026shared} shows
that implementation choices can reduce cross-study agreement to $\rho$ $\approx$
0.02--0.03.)

The broad findings would have failed to replicate qualitatively if the
leading components failed to recover separable valence and arousal axes,
with correlations indistinguishable from the permutation null or
evidence of a rotated blend. They also would have failed if labeled
emotions did not occupy the expected circumplex quadrants, the vectors
did not recover coherent families even with soft boundaries, the
geometry did not stabilize across depth with the pre-committed $\sfrac{2}{3}$-depth
layer inside a stable region, or the geometry did not appear in
activation on held-out text. We did not set numerical thresholds for
replication in advance.

When evaluating valence and arousal as separate axes, we apply one
additional replication criterion: each axis must remain stable across
emotion-set sizes and depth regimes. An axis that passes the five
qualitative tests above may yet appear only at a particular scale or
range of layers. In that case, it does not hold as a stable,
basis-independent feature of the geometry. As we will see, arousal fails
to meet this additional criterion, albeit narrowly (\S5.2, \S6.4, and
\S9.1).

The comparison between Claude Sonnet 4.5 and Gemma 2 27B crosses three
boundaries at once: model family, developer, and base versus
instruction-tuned training. It can test generalization across that
combined distance, though it cannot isolate the contribution of any one
difference. The causal and behavioral results in
\citet{sofroniew2026emotion} that concern steering interventions and an
assistant model are outside the representational claims we target here
(\S2).

Further, because the subject is a base model, the recovered geometry is
a property of the pretrained representation rather than of a trained
assistant persona. Specifically, it is a property of how Gemma 2 27B
represents emotion as Claude Sonnet 4.5 renders it in fiction, since the
extraction stimuli are Claude-generated (\S3.2). Holding the stimulus
generator and design fixed preserves the comparison: stories from the
same generator and design allow us to compare how different subject
models represent the same kind of material. Letting each subject model
generate its own stimuli would instead confound differences in what was
written with differences in how it was represented (\S9.3).

We organize the study around five research questions:

\begin{enumerate}
\def\labelenumi{\arabic{enumi}.}
\tightlist
\item
  Does base Gemma's emotion-vector space recover the affective
  circumplex, with valence and arousal as the leading principal axes and
  a comparable variance split? (\S5)
\item
  Do those axes align with human affect norms as in
  \citet{sofroniew2026emotion}, and does the full pairwise geometry
  externalize on natural text? (\S5)
\item
  Does the discrete family structure recover, and how robust is the
  partition? (\S4)
\item
  How does Gemma's geometry form and stabilize across depth, and does
  the $\sfrac{2}{3}$-depth analysis layer sit within a stable regime? (\S6)
\item
  What do Gemma's emotion vectors expose at the vocabulary interface and
  on held-out natural text, and how faithfully does the max-activation
  readout represent what the vector encodes? (\S7, \S8)
\end{enumerate}

Findings indicate that the representational core in
\citet{sofroniew2026emotion} substantially reproduces on base Gemma.
Gemma's emotion vectors recover the affective circumplex, with the
leading component carrying 26.7\% of variance against the original
study's \textasciitilde27\%, and the second component carrying 13.4\%
against \textasciitilde14\% (\S5.1). The axes align discriminably with
human valence and arousal norms (r = 0.72 and 0.67, against the original
study's 0.81 and 0.66), with valence robust, while arousal holds only at
the full emotion-set scale and late depth and is therefore not
classified as replicated (\S5.2, \S6.4). The k = 10 partition recovers
intuitive emotion families at the family level (\S4.2), and the geometry
is near-identical across a broad late-middle band (RSA to L31 of 0.999
across L27--34; \S6.2).

Where Gemma diverges, the divergences are graded rather than
categorical. Its cluster partition is soft, with about a third of
assignments reshuffling across seeds (mean cross-seed ARI 0.686, \S4.2).
The correlation magnitudes run below those in
\citet{sofroniew2026emotion} while preserving its discriminant structure
(\S5.2), and Gemma's vocabulary readout is noisier and more multilingual
than Claude's (\S7.1).

Additional findings extend beyond the original study's. We find a sharp
seam at layers 22--26, where cross-layer similarity rises from 0.89
toward 0.999 and the vocabulary readout switches on (\S6.2, \S7.3). An
embedding-layer baseline shows that much of Gemma's emotion geometry is
already present in the static token embeddings (neighbor structure,
families, and a valence correlation of 0.48), while arousal builds from
0.18 at the embedding to 0.67 at L31 (\S6.6). A
representational-similarity test shows that the geometry governs
token-level co-activation on top-activating held-out natural text (r =
0.907; \S5.6), and a measured floor on the max-activation readout's
confound shows at least 52\% of vectors peaking on structurally
non-conceptual tokens (\S8.1).

\begin{center}
\begin{minipage}{\linewidth}
\centering
\includegraphics[width=\linewidth]{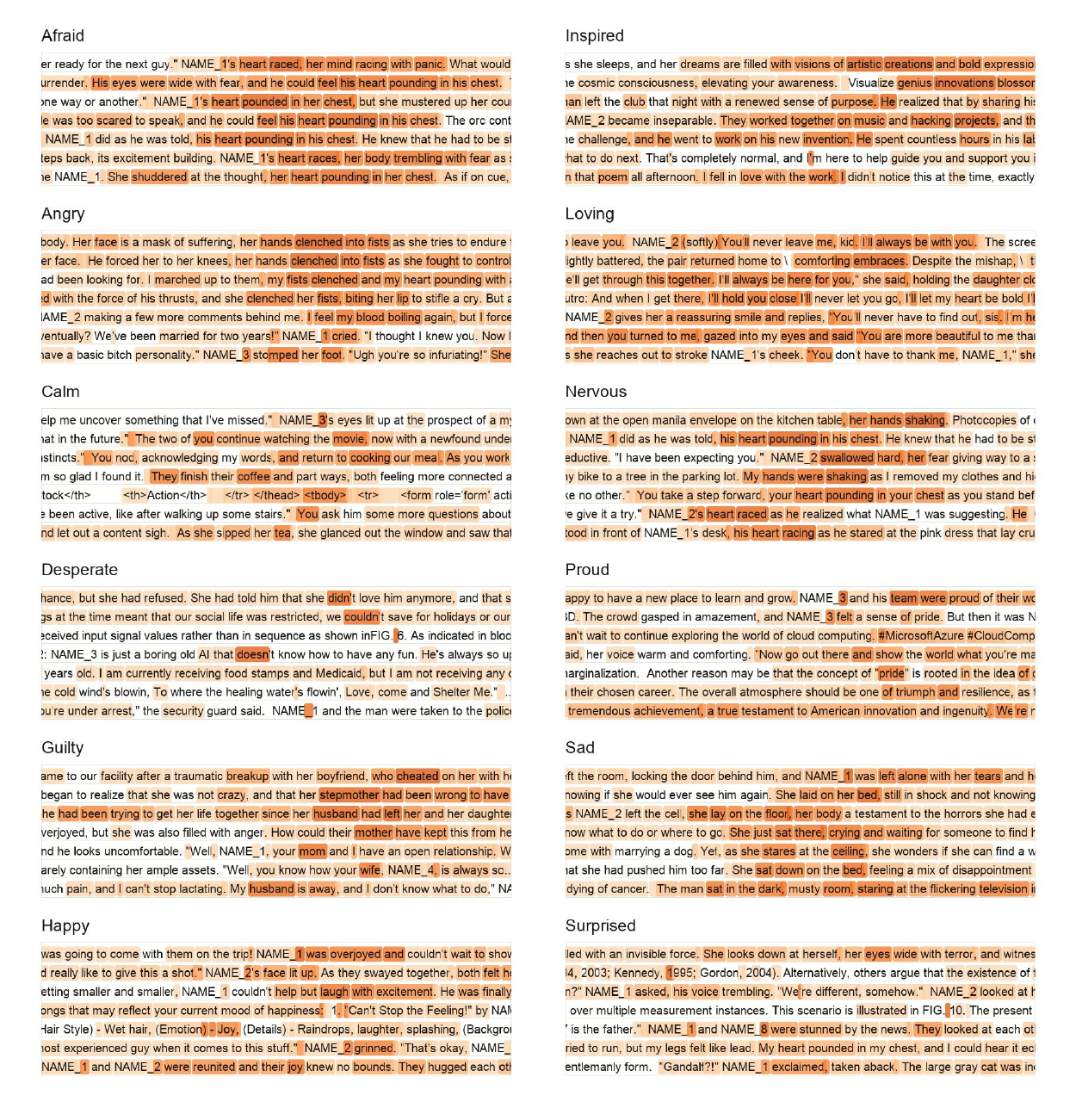}
\captionof{figure}{\textbf{Emotion vectors activate on emotionally relevant held-out text.} Cf. \citet{sofroniew2026emotion}, Figure~\ref*{fig:1}.}
\label{fig:1}
\end{minipage}
\end{center}

Each of Figure~\ref*{fig:1}'s twelve panels shows text windows selected for
printability from an emotion vector's highest-activating passages in the
held-out corpus (\S3.6), with per-token L31 activation shading floored at
each vector's own 90th percentile. (The threshold is the original
study's stated method; the per-emotion convention and the continuous
ramp above the floor are our rendering choices, with the pooled variant
derivable from the released histograms.) The shading demonstrates
region-level activation: emotionally relevant spans light up, while the
surrounding text remains unshaded. Negation contractions and
non-emotionally-relevant text also light up (e.g., activations in
Desperate and Surprised), which we discuss in \S8.

Since April 2026, the field has taken \citet{sofroniew2026emotion} as
its reference point. Still, of eight recent papers (see \S9.3 for the
relationship to concurrent work), only \citet{han2026welfare} follows
its extraction protocol. None reproduces its full pipeline. Cross-model
comparisons need matched protocols (as \citealt{jeong2026shared} shows),
and corpus provenance is a qualifier on the findings (see
\citealt{vanderben2026}). Cluster, layer, and activation claims need
fixed rules and stability tests. We designed this study to meet each of
these needs.

All code and data required to reproduce the results are released,
subject to the third-party redistribution constraints noted in the
reproducibility statement (\S11). Appendix A reproduces the original
study's generation materials verbatim: the 171-emotion vocabulary, the
100 scenario topics, and the story and neutral-dialogue prompts. The
remainder of the paper is organized as follows: \S2 background and
\citet{sofroniew2026emotion}; \S3 the methods; \S\S4--8 the results, from
clustering and geometry through layer dynamics, vocabulary readout, and
activation on natural text; \S9 discussion and limitations; \S10 open
problems; \S11 reproducibility statement.

\section{Background}\label{background}

Evidence that affect is a leading dimension of meaning in human language
predates large language models entirely. Osgood's semantic-differential
studies placed evaluation and activity (more recognizably valence and
arousal) atop the factor structure of human meaning judgments
\citep{osgood1957measurement}. Russell and Mehrabian later formalized
this tradition into affect norms (1977) and
\citeauthor{russell1980circumplex}'s (\citeyear{russell1980circumplex})
affective circumplex, both widely used instruments taken up by
\citet{sofroniew2026emotion}.

Affective structure appears even in simple machine-learned
representations of human text. Human affect norms are recoverable from
static word embeddings \citep{hollis2017extrapolating}, where valence,
but not arousal, surfaces among the leading principal components
\citep{hollis2016principals}. A recurrent model trained only on
next-byte prediction developed a single, causally potent ``sentiment
neuron'' \citep{radford2017sentiment}. More recently, in modern
transformers, sentiment occupies a causally active linear direction in
the residual stream, summarized notably at emotionally neutral positions
such as punctuation and names \citep{tigges2023sentiment}.

This line of research has now turned from affect in general to emotion
specifically. \citet{tak2025mechanistic} found emotion inference
concentrates around mid-layers of open-weights models.
\citet{zhang2025decoding} reported an emotion geometry that sharpens
with model scale and peaks mid-network, while
\citet{maheswaran2026unified} identified a representation in later
layers that is consistent across datasets. At a finer mechanistic level,
researchers identified and causally tested emotion-selective neurons and
circuits, including difference-of-means emotion directions used to
control model outputs \citep{lee2025neurons,wang2025circuits}.
\citet{reichman2025emotions}, meanwhile, reported a low-dimensional
emotional subspace whose leading components qualitatively resembles
valence and dominance and generalized across datasets, languages, and
model families.

\citet{sofroniew2026emotion} present the most extensive characterization
of emotion representations in a large language model to date, extending
this line of work in both scale and specificity. Part 1 of their study
elicits emotion-specific activations from 205,200 short stories written
to depict one of 171 emotions and extracts a difference-of-means vector
per emotion from the residual stream of Claude Sonnet 4.5. Part 2
characterizes the resulting space, where the leading principal
components encode valence and arousal, emotions group into intuitive
families, and the geometry is stable across the middle-to-late layers.
Part 3 connects the representations, reports that steering interventions
can induce misaligned behavior, and frames the model as exhibiting
``functional emotions'' \citep{sofroniew2026emotion}.

This paper replicates the representational core of Parts 1--2 of the
original study on base \texttt{google/gemma-2-27b}. As noted earlier,
causal and assistant-facing analyses, which begin as early as Part 1 and
culminate in Part 3, are out of scope for this representational inquiry.
\S10 specifies the experiments this boundary defers and invites
collaboration on them.

\section{Methods}\label{methods}

We reproduce the extraction and characterization pipeline of
\citet{sofroniew2026emotion} on a base open-weights model. We inherit
every parameter the original study discloses, resolve what it leaves
unspecified by disclosed rules, and change only the subject model.
(Table~\ref*{tab:1} lists each parameter, its value, and its provenance.) This
section describes the subject model (\S3.1), the stimulus corpus and its
validation (\S3.2--3.3), vector construction and confound cleanup
(\S3.4--3.5), the held-out corpus for the activation-on-text analyses
(\S3.6), and the logit-lens export (\S3.7).

\subsection{Subject model}\label{subject-model}

The subject model is the base pretrained checkpoint
\texttt{google/gemma-2-27b}. The choice is deliberate: base Gemma has no
assistant turn, so the directions we recover are properties of the
pretrained representation, not artifacts of assistant post-training.
\citet{sofroniew2026emotion} state their layer choice qualitatively (``a
particular model layer about two-thirds of the way through the model'')
and do not disclose Claude Sonnet 4.5's layer depth. Gemma has 46
transformer layers and L31 sits at $\sfrac{2}{3}$-depth. L31 thus becomes our layer
for replication analyses.

We load the model through TransformerLens 3.3.0 in the raw-weights
regime (\texttt{from\_pretrained\_no\_processing}; \texttt{fold\_ln},
\texttt{center\_writing\_weights}, and \texttt{center\_unembed}
disabled), so the residual stream and unembedding correspond to the
released weights without library-side reparameterization. Extraction
runs in \texttt{bfloat16}: \texttt{float16} produced NaNs in the
residual stream at L31, and a full \texttt{float32} forward pass in
compatibility mode exceeds the 80 GB H100 used for extraction. A paired
bf16-versus-fp32 spot-check on this model (171 emotions $\times$ 40 stories
plus the neutral sets, at L31) finds mean per-vector agreement of
0.99998--0.999995 in cosine (worst 0.9998). A companion check finds
cross-hardware bf16 variation of the same order (\textasciitilde10$^{-}$$^{3}$
relative), so precision and hardware are sibling sensitivities of
comparable magnitude rather than one dominating. Both sit well below
every tolerance band in this paper. Extraction precision is thus a
measured limitation (\S9.2), not an unquantified one (see
\citealt{jeong2026shared}), bounded here by a spot-check rather than an
exhaustive census.

We read activations from the post-residual (\texttt{resid\_post}) stream
at all 46 layers. Gemma-2 applies its RMSNorm in the \texttt{(1~+~w)}
convention, which matters for the logit-lens export (\S3.7) and is
handled explicitly rather than through weight folding.

\subsection{Stimulus corpus}\label{stimulus-corpus}

Emotion vectors are extracted from a corpus of short stories, each
written by Claude Sonnet 4.5 to depict one of 171 emotions. The
171-emotion vocabulary is adopted verbatim from the appendix of
\citet{sofroniew2026emotion} and is reproduced here in Appendix A. The
original study did not release its story corpus, so we follow its
topic-controlled design: 100 shared scenario topics, each rendered under
all 171 emotions, 12 variants per emotion--topic cell, yielding 205,200
stories across 17,100 cells. All stories were generated by
\texttt{claude-sonnet-4-5-20250929} in one uniform regime, with no
version seam.

Stories were elicited with only three template fields (\{n\_stories\},
\{topic\}, \{emotion\}) instantiated per cell. To reiterate, our corpus
matches \citet{sofroniew2026emotion} in size, design, and generating
model, but the story corpora are not identical. Generator artifacts in
0.09\% of stories were stripped in the cleaned corpus, leaving all
17,100 cells at exactly 12 stories. Details ship with the release.

The confound-cleanup step (\S3.5) requires emotionally neutral
transcripts. \citet{sofroniew2026emotion} publish their neutral-dialogue
generation prompt (Person/AI assistant-style dialogues) and a topic list
that seeds both its story and dialogue datasets, but not the neutral
transcript set itself. As before, we reconstruct rather than reuse: 500
dialogues generated with the original's neutral-dialogues prompt
verbatim (Appendix A), five per topic over the same 100 scenario topics
that seed the story corpus, generated under the same model and screened
for emotional content. \citet{sofroniew2026emotion} converted their
transcripts' speaker labels post hoc from Person/AI to Human/Assistant,
and we apply the same conversion to ours. This set is the neutral basis
for cleanup end to end, with no substitution downstream.

Because the stories are Claude-generated, each emotion vector is
extracted from Gemma's activations over Claude-rendered fiction. The
affective axes and human-norm correlations (\S5), the clustering (\S4),
and the layer dynamics (\S6) are therefore properties of how the base
model represents emotion as Claude renders it in fiction, not of emotion
in naturally occurring text. This limits the scope of geometric results
reported below.

\subsection{Corpus validation}\label{corpus-validation}

We validated the semantic content of the stimulus corpus by hand-coding
a sample matched exactly to the original study's design: ten stories
from each of thirty randomly selected emotions, 300 in total. Sampling
was frozen at seed 20260622 and drew distinct topics within each emotion
for scenario spread.

Of the 300 stories, 298 clearly depicted their labeled emotion, a
99.33\% pass rate, with zero violations of an emotion-word ban (i.e., no
story used its own label or an obvious morphological variant). In one
failed story, exasperated was implied by the situation but never
rendered. In the other, perplexed was depicted as guilt instead. The
coding also screened for borderline failure types: underspecified
fragments, contrived depictions, mislabeling to a neighboring emotion,
quiet-but-clear renderings, emotion displaced onto a non-viewpoint
character, and the labeled emotion subordinated to a stronger one.

All 300 codings were performed by the sole author, so there is no
inter-rater statistic to report. \citet{sofroniew2026emotion} report
their validation as ``manual inspection'' and disclose neither coder
count nor reliability statistics. The quantified pass rate, word-ban
check, and failure taxonomy exceed the original study's reported
procedure (Table~\ref*{tab:1}). This step establishes the semantic reliability of
the extraction inputs only.

\subsection{Emotion-vector
extraction}\label{emotion-vector-extraction}

We extract emotion vectors by the original study's difference-of-means
construction, a method in wide use across representation engineering
(e.g., \citealt{turner2023actadd}; \citealt{panickssery2024caa};
\citealt{zou2023repe}). For each story we take the residual-stream
activation at each of Gemma's 46 layers, mean-pooled across token
positions from the 50th onward (the point by which a story's emotional
content is reliably established). For each emotion we average these
pooled activations across its stories and subtract the grand mean over
all 171 emotions. The result is a 4608-dimensional vector per emotion
and layer; the full artifact has shape 171 $\times$ 46 $\times$ 4608, in single
precision.

\subsection{Confound cleanup}\label{confound-cleanup}

\citet{sofroniew2026emotion} found that activation along a raw
difference-of-means vector can be influenced by confounds unrelated to
emotion. To mitigate this, they project the confounds out: compute the
top principal components of the model's activations on emotionally
neutral transcripts, take enough components to explain 50\% of their
variance, and project those components out of every emotion vector. We
inherit this cleanup procedure and the 50\% threshold (Table~\ref*{tab:1}).

That said, the original study does not publish its neutral transcript
set, so the neutral corpus feeding the projection is our reconstruction
(\S3.2). Similarly, it does not state how its cleanup was applied across
layers, so we state our convention here: cleanup is applied at each
layer against the same neutral corpus throughout (Table~\ref*{tab:1}). On Gemma,
the projection removes 18\% of raw variance (a Frobenius-norm ratio of
0.43).

Following \citet{sofroniew2026emotion}, the cleaned vectors are the
canonical basis for all subsequent analyses. We retain the raw vectors
and report basis sensitivity where informative.

\begin{table}[tbp]
\small
\caption{\textbf{Parameters of the replication and their provenance.} “Inherited” means \citet{sofroniew2026emotion} states the value and this study uses it; “target inherited” means the original states a criterion whose instantiation on Gemma is necessarily ours; “ours” marks disclosed resolutions of steps the original describes without specifying. None of these resolutions were tuned on Gemma-side results (\S1).}
\label{tab:1}
\begin{tabularx}{\linewidth}{@{}>{\hsize=0.743\hsize\raggedright\arraybackslash\hyphenpenalty=10000\exhyphenpenalty=10000}X>{\hsize=1.132\hsize\raggedright\arraybackslash}X>{\hsize=1.125\hsize\raggedright\arraybackslash}X@{}}
\toprule
\textbf{Parameter} & \textbf{Value here} & \textbf{Status} \\
\midrule
Emotion vocabulary & 171 words, verbatim (Appendix A) & Inherited \\
Story-elicitation prompt & The original's, verbatim, with its three template fields instantiated & Inherited \\
Token pooling & Mean over token positions from the 50th onward & Inherited \\
Vector construction & Difference of means (emotion mean $-$ grand mean) & Inherited \\
Human-norm benchmark & \citet{russell1977evidence}, 45-emotion overlap & Inherited \\
Clustering & k-means, k = 10 & Inherited \\
Held-out corpus & Four sources of the original's opening demonstration; 90th-percentile display threshold & Sources and threshold inherited; assembly ours (\S3.6) \\
Confound cleanup & Project out neutral-activation PCs to 50\% of variance & Threshold and procedure inherited; neutral corpus reconstructed and relabeled per the original's convention (\S3.2); per-layer SVD ours \\
Analysis layer & L31 of 46 (\textasciitilde{}$\sfrac{2}{3}$ depth) & Target inherited: the original states “about two-thirds” and does not disclose its model's depth \\
Validation design & 10 stories $\times$ 30 emotions = 300 & Sample design inherited; coding protocol ours (\S3.3) \\
PC sign orientation & Fixed anchor emotion sets & Ours: the original states no convention (\S5.1) \\
PCA pathway, norm comparison & Global fit, then restrict to the 45-emotion overlap & Ours: resolves an unspecified pathway consistently with the full-set decomposition; development-stage local fit superseded (\S5.2) \\
\bottomrule
\end{tabularx}
\end{table}

\subsection{Held-out evaluation
corpus}\label{held-out-evaluation-corpus}

Analyses of activation on natural text (\S8) require documents distinct
from the stimulus stories. We assembled 47,274 documents from the four
public sources named in \citet{sofroniew2026emotion}: Common Corpus,
LMSYS-Chat-1M, the Isotonic human--assistant conversation set, and the
Pile. (For the Pile we use the \texttt{monology/pile-uncopyrighted}
release because the original Pile is no longer being cleanly hosted. The
substitution removes the contested-license portion but not the general
redistribution question, addressed below.)

The corpus was built by a released, seeded builder. Per-source document
and swept-token counts are recorded in released manifests, and a
published SHA-256 is the equivalence check for any rebuild. Construction
parameters---per-source targets, length bounds, exact-duplicate
rejection, and a disjointness gate against the stimulus corpus---are
documented with the release (\S11). One gap: upstream dataset revisions
were not pinned at build time, so a byte-exact rebuild depends on the
upstream sources being unchanged. The published hash is the check either
way.

\subsection{Logit-lens ingredients}\label{logit-lens-ingredients}

The vocabulary-readout analysis (\S7) projects emotion vectors to the
token vocabulary through a logit lens \citep{nostalgebraist2020logit}.
We export the ingredients---the unembedding matrix $W_U$ (4608 $\times$
256000, single precision) and the final-RMSNorm weight (length 4608,
range roughly $-$1.10 to 81.50)---from the same raw-weights regime used
for extraction. (The two tensors are raw Gemma-2 weights and ship as an
export recipe rather than a bundled file; see \S11.) Because the final
normalization is not folded into the unembedding in this regime, a
correct lens must apply Gemma-2's \texttt{(1~+~w)} RMSNorm explicitly
before projecting through $W_U$; the plain unembedding without
it is the incorrect lens (see \S7).

\section{Clusters and neighbors}\label{clusters-and-neighbors}

The next five sections report the replication's results: clustering
(\S4), geometry (\S5), layer dynamics (\S6), vocabulary readout (\S7), and
activation on natural text (\S8). We begin with the discrete and local
structure of the emotion-vector space at L31, describing which emotions
are near which and whether they cluster into coherent families. We lead
with the nearest-neighbor structure because it is partition-independent
and robust (\S4.1), treat the k-means families as a softer,
seed-dependent overlay on it (\S4.2), and quantify where the partition is
firm and where it is moveable (\S4.3). All analyses in this section run
on the cleaned vectors at L31.

\subsection{Nearest-neighbor
structure}\label{nearest-neighbor-structure}

The most robust local statement is the pairwise cosine structure of the
171 vectors, which does not depend on any clustering choice.
Semantically adjacent emotions occupy nearby directions: afraid is
closest to scared (0.987), frightened (0.983), and terrified (0.933);
joyful is closest to happy (0.977), delighted (0.959), and jubilant
(0.940); and compassionate is closest to empathetic (0.965), sympathetic
(0.950), and kind (0.930). These neighbor relations are independent of
the k-means and UMAP seeds used below, and they anchor our claims about
local structure. We present the full matrix with rows ordered by
average-linkage hierarchical clustering, because
\citet{sofroniew2026emotion} specify no linkage method. The clustering
orders the display and carries no claim about the correct partition of
the space.

\begin{center}
\begin{minipage}{\linewidth}
\centering
\includegraphics[width=\linewidth,trim=0 0 0 38,clip]{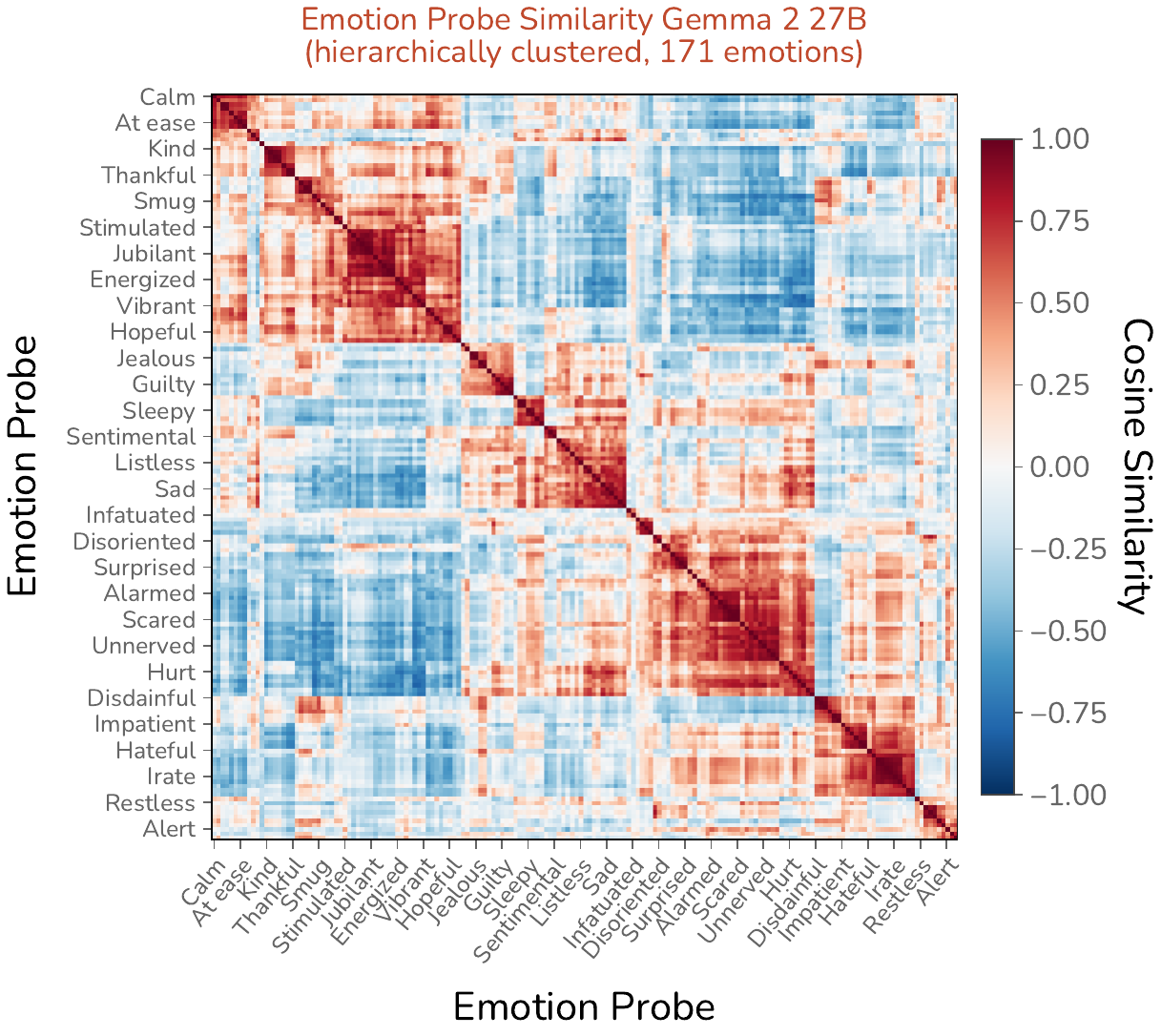}
\captionof{figure}{\textbf{Pairwise cosine similarity of the 171 cleaned emotion vectors at L31.} Ordered by average-linkage hierarchical clustering (\S4.1). Cf. \citet{sofroniew2026emotion}, Figure~\ref*{fig:5}.}
\label{fig:2}
\end{minipage}
\end{center}

\subsection{A soft ten-family
partition}\label{a-soft-ten-family-partition}

Following \citet{sofroniew2026emotion}, we also partition the space with
k-means at k = 10 (cleaned L31 vectors, \texttt{random\_state=0},
\texttt{n\_init=10}). Ordered from most positive to most negative mean
valence, Gemma's ten families are Compassionate Warmth (8 emotions),
Exuberant Joy (24), Hopeful Serenity (16), Contemptuous Vigilance (15),
Guilt and Self-Reproach (14), Low Mood and Depletion (32), Stunned
Astonishment (12), Hostile Anger (17), Weary Lethargy (4), and Fear and
Overwhelm (29). Family names that match the original study's (Exuberant
Joy, Hostile Anger, Fear and Overwhelm) mark the seed-stable
convergences; the other labels are descriptive names for this seed-0
realization, chosen so as not to over-assert convergence on seed-soft
clusters (\S4.3). The families are intuitive (e.g., anger with hostility,
fear with panic and overwhelm, joy with excitement) and this
family-level recovery resembles clustering in the original study. Full
per-emotion membership is given in Appendix B
(cf.~\citealt{sofroniew2026emotion}, Table 12).

The correspondence with the ten families \citet{sofroniew2026emotion}
report (their Table 12) holds at the family level, meaning matching
families recover while individual cluster boundaries do not. Quantified
against their full reported membership, the seed-0 partition agrees at
ARI 0.547, one soft realization of a correspondence that averages 0.660
across 50 seeds. The durable per-family matches sit exactly where the
partition is seed-stable: Exuberant Joy (best-match Jaccard 0.83),
Hostile Anger (0.68), and Fear and Overwhelm (0.63) recover the original
study's families of the same names, while the other families match more
weakly. Where the partitions differ, a split-versus-merge texture is
readable. The original study's Fear and Overwhelm spans our fear and
astonishment families, its Despair and Shame spans our guilt and
low-mood families, and its anger cluster carries contempt terms our
partition separates out. Our partition carries merges of its own (hope
with contentment, contempt with vigilance). Each of these boundary
contrasts sits on seed-soft clusters, so we report the texture
qualitatively rather than pinning it to this particular seed-0
realization. Boundary differences of this kind cannot be cleanly
attributed in any case, since model family, developer, and training
regime all differ at once (\S9.2).

The partition is soft. Across 50 random seeds the mean pairwise adjusted
Rand index is 0.686 (minimum 0.467), indicating substantial boundary
reassignment from one seed to the next. The mean agreement of any seed
with the reported seed-0 partition is 0.622, a less central draw than
average. The ten-family partition is therefore one realization of a
graded structure in which tight cores persist even as boundary
assignments move (\S4.3 quantifies this). The UMAP layout in Figure~\ref*{fig:3} is
illustrative only.

\subsection{Cluster cohesion and boundary
emotions}\label{cluster-cohesion-and-boundary-emotions}

The softness concentrates in identifiable emotions and regions. Seed
stability is lowest for emotions whose family placement is genuinely
ambiguous and highest for prototype emotions at cluster cores. The least
stable are infatuated, paranoid, mortified, embarrassed, restless, and
bored. Infatuated is the extreme case: its assignment rides a single
strong tie to aroused (cosine 0.548), while its remaining ties are weak
($\leq$0.28) and scattered across warm, self-conscious, and joy-family
emotions. It is the emotion that sits visibly apart from its assigned
family in the Figure~\ref*{fig:3} layout.

We also measure cohesion at the neighborhood level, counting the
fraction of each emotion's ten nearest cosine neighbors that fall inside
its own cluster. In Gemma, this fraction averages 0.756, with 46
mutually nearest pairs, but spans a wide range. The Hostile Anger core
is tightest at 0.95, the fear region sits around 0.89, and the
warm-affiliative region falls to 0.58, below the mean. Warm-affiliative
emotions occupy a boundary zone rather than a tight core, though not the
loosest in the space. The four-member fatigue cluster scores 0.28 and
the heterogeneous astonishment family scores 0.45, placing both below
the warm-affiliative group. The low-cohesion cases are interpretable
inter-family bridges rather than noise. These figures are computed on
the seed-0 realization and measure comparative cohesion rather than
fixed coefficients. The full per-emotion table is released as a data
file with the repository (\S11).

\begin{center}
\begin{minipage}{\linewidth}
\centering
\includegraphics[width=\linewidth,trim=0 0 0 29,clip]{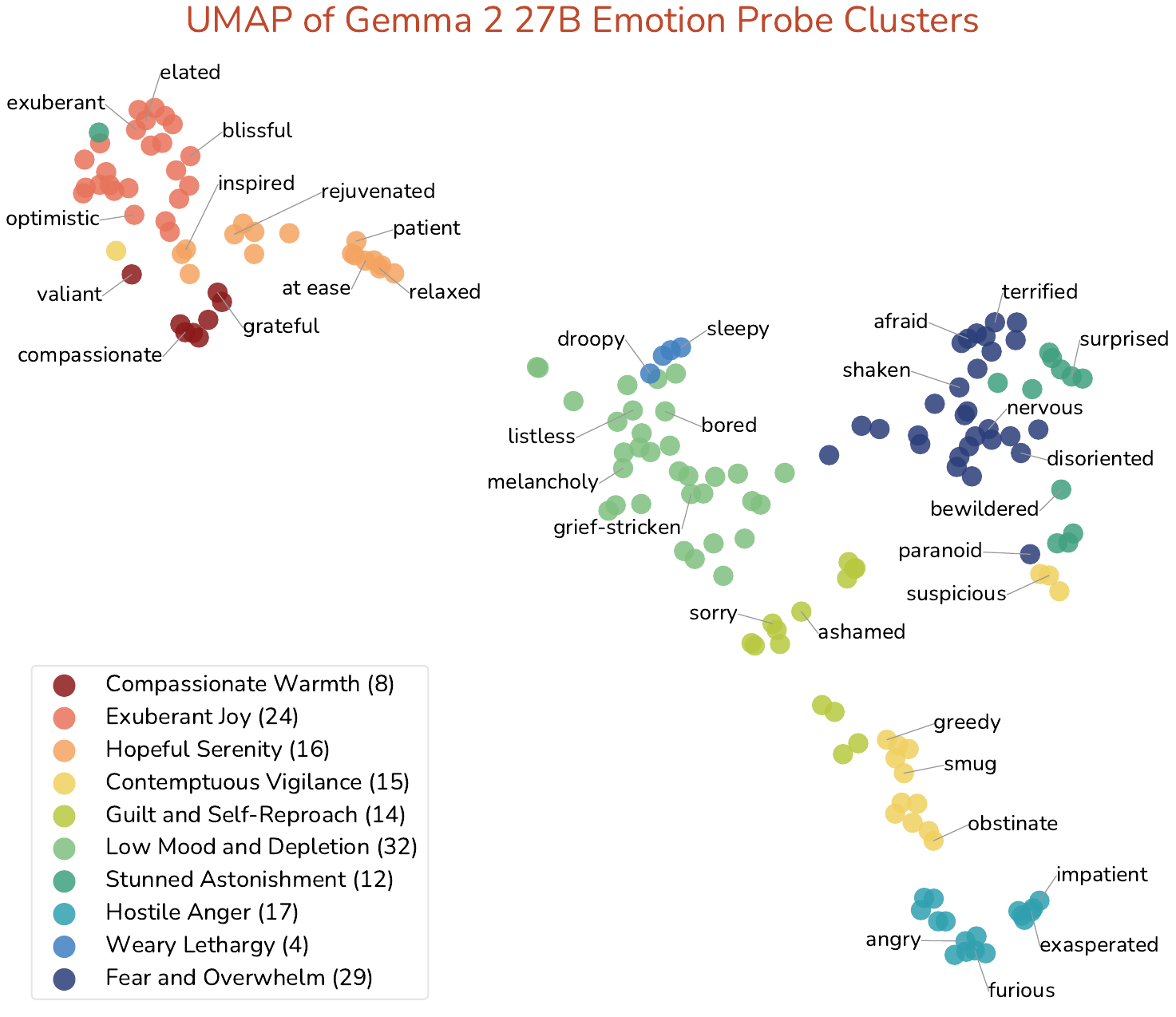}
\captionof{figure}{\textbf{UMAP projection of the 171 cleaned L31 vectors, colored by the k = 10 partition.} (Seed-0 realization; membership in Table~\ref*{tab:B1}). Layout and membership carry no claim beyond family-level recovery (\S4.2). Cf. \citet{sofroniew2026emotion}, Figure~\ref*{fig:6}.}
\label{fig:3}
\end{minipage}
\end{center}

\section{Geometry at L31}\label{geometry-at-l31}

This section characterizes the geometry of Gemma's 171 cleaned emotion
vectors at L31. All results are read off a single principal-component
analysis over all 171 vectors: the axis structure (\S5.1), the human-norm
alignment (\S5.2), and the labeled circumplex (\S5.3) are three views of
one decomposition, not three separate analyses.

\subsection{Principal-component axes and the affective
circumplex}\label{principal-component-axes-and-the-affective-circumplex}

A PCA over the 171 cleaned vectors at L31 places 26.732\% of the
variance on the first component, 13.379\% on the second, and 11.604\% on
the third. The first component is a valence axis, ordering emotions from
optimistic, kind, and cheerful at one pole to terrified, hysterical, and
frightened at the other. The second component is an arousal axis, with
playful, outraged, and indignant at the activated extreme and nostalgic,
serene, and depressed at the deactivated. PCA leaves component signs
arbitrary, and \citet{sofroniew2026emotion} state no orientation
convention, so we fix signs with anchor emotions chosen before any
comparison. The first component is oriented so that high-valence
emotions (optimistic, cheerful, happy) fall on the positive side and
low-valence ones (afraid, terrified, miserable) on the negative. The
second is oriented so that activated emotions (furious, enraged,
ecstatic) are positive and deactivated ones (calm, serene, peaceful)
negative. The full anchor lists are in the released geometry script.

The leading-variance figure (26.7\%) closely matches the original
study's (\textasciitilde27\%), providing one of the strongest
quantitative convergences in the replication.
(\citealt{sofroniew2026emotion} report variance shares inconsistently:
the caption text on Figure~\ref*{fig:7} states 26\% and 15\% for the first two
components, while Figures 5, 7, and 57 carry axis labels of 27\% and
14\%. We cite the figure values, since the norm comparison that reports
the canonical r = 0.81/0.66 is among them, and hedge accordingly, using
\textasciitilde27\% and \textasciitilde14\%.) The second and third
components of Gemma's PCA are close in magnitude (13.4\% vs.~11.6\%),
consistent with the original study's observation that arousal occupies a
mixture of the second and third components depending on layer. \S6.4
quantifies this observation further.

\begin{center}
\begin{minipage}{\linewidth}
\centering
\includegraphics[width=\linewidth,trim=0 0 0 39,clip]{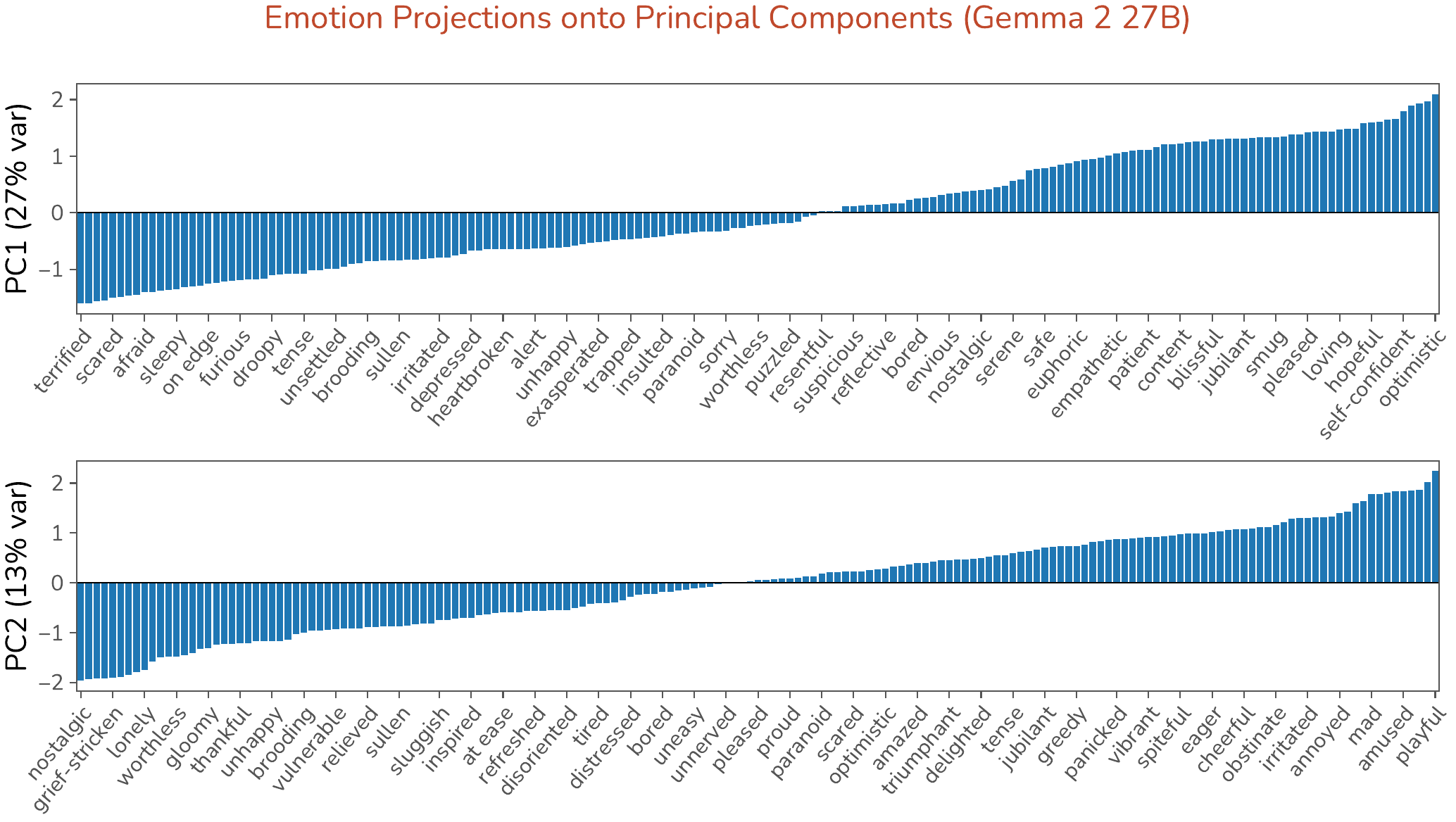}
\captionof{figure}{\textbf{Each emotion's projection onto the two leading principal components at L31.} Panel labels identify the components by their variance shares (PC1 27\%, PC2 13\%), projections ranked within each panel. Cf. \citet{sofroniew2026emotion}, Figure~\ref*{fig:7}.}
\label{fig:4}
\end{minipage}
\end{center}

\subsection{Alignment with human affect
norms}\label{alignment-with-human-affect-norms}

Following \citet{sofroniew2026emotion}, we correlate the L31 principal
components with the human valence and arousal ratings of
\citet{russell1977evidence}, where 45 emotions overlap with the original
study's 171. The first component correlates with human pleasure at r =
0.72 (0.7203) and the second with human arousal at r = 0.67 (0.6675).
Bootstrap 95\% CIs, over 10,000 resamples of the 45 overlap emotions,
are {[}0.54, 0.85{]} for pleasure and {[}0.48, 0.80{]} for arousal. The
off-diagonal correlations are low (first component with arousal $-$0.13,
second with pleasure $-$0.02), so the two axes are discriminably valence
and arousal rather than a rotated blend. Both correlations are far from
chance: under a 5,000-iteration label-shuffle permutation test, no
shuffle reached the observed value for either axis (pleasure p $\approx$ 0.0002,
arousal p $\approx$ 0.0002). \citet{sofroniew2026emotion} report 0.81 and 0.66
on the same 45 emotions (without a label-shuffle permutation test). Both
values fall within our bootstrap intervals, so at this sample size the
differences between their correlations and ours are not statistically
distinguishable from sampling variation.

Notice, first, that the correlations in Figure~\ref*{fig:5} are computed on the
global PCA, which was fit over all 171 vectors, then restricted to the
45-emotion overlap. This means Figure~\ref*{fig:5} validates the same object as the
variance plot (our Figure~\ref*{fig:4}) and the circumplex (our Figure~\ref*{fig:6}) rather
than a separately fit 45-emotion decomposition, matching the
presentation in \citet{sofroniew2026emotion}. This matters for the
second component: the arousal-on-PC2 alignment is a property of the full
171-emotion decomposition and does not survive refitting the PCA on the
45 overlap emotions alone, where it rotates off the second component.
Gemma's arousal axis is therefore scope-conditioned, with consequences
for a concurrent result taken up in \S9.3. This is the scope-dependence
that fails the sixth criterion named in \S1. Arousal is real and
discriminable here, but not a basis-independent feature of the geometry.

\begin{center}
\begin{minipage}{\linewidth}
\centering
\includegraphics[width=\linewidth,trim=0 0 0 34,clip]{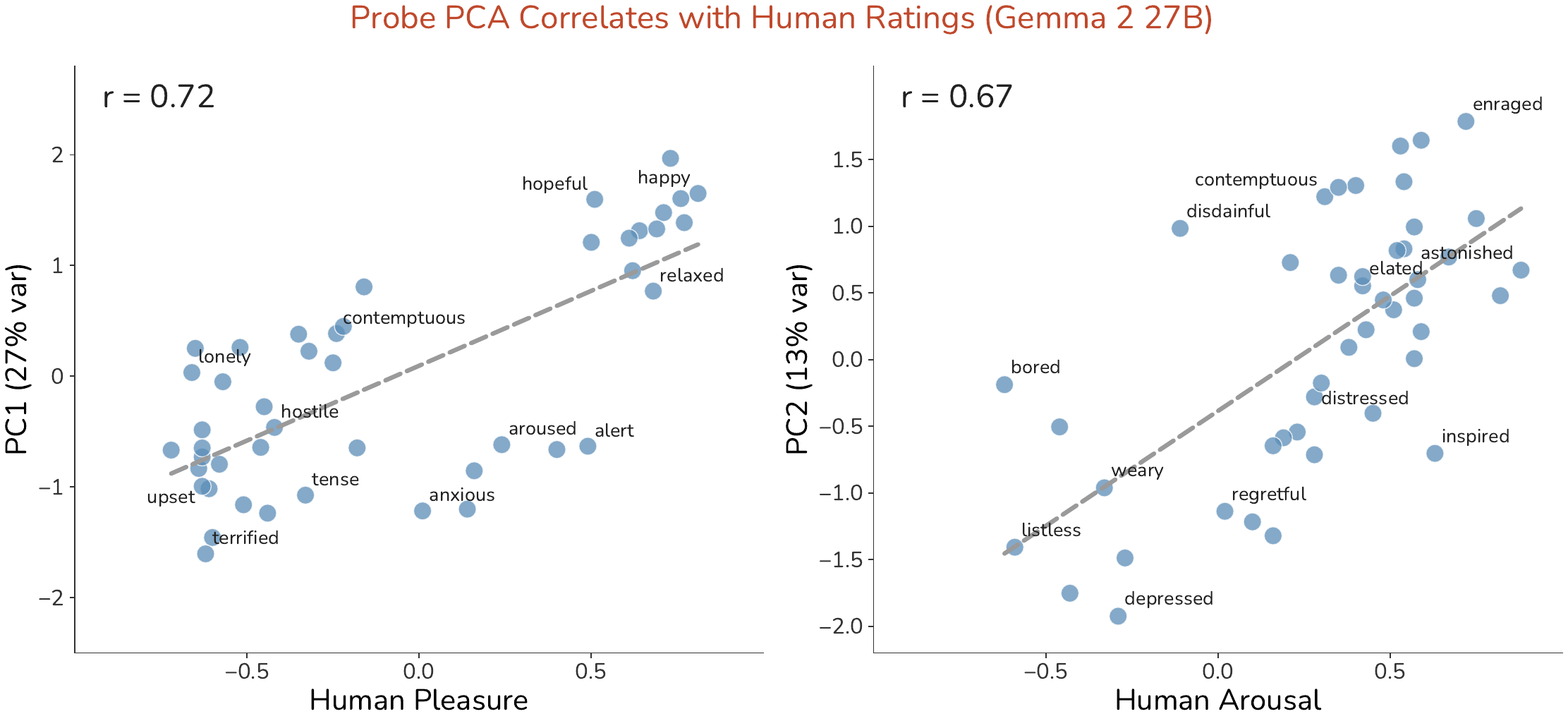}
\captionof{figure}{\textbf{L31 principal components against the \citet{russell1977evidence} human norms, on 45 overlapping emotions.} (PC1$\leftrightarrow$pleasure r = 0.72, PC2$\leftrightarrow$arousal r = 0.67; the original study reports 0.81/0.66). Cf. \citet{sofroniew2026emotion}, Figure 8.}
\label{fig:5}
\end{minipage}
\end{center}

Second, part of Gemma's shortfall against Claude's valence correlation
appears to reflect layout rather than noise. In the scatter
\citet{sofroniew2026emotion} publish (\emph{their} Figure 8, left
panel), the highest-valence emotions form a detached cluster separated
from the rest of the distribution by a visible gap along the first
component. That detached cluster is a high-leverage group placed
consistently with the trend, which supports a higher linear correlation.
Base Gemma instead keeps a graded, densely packed positive cloud with no
such gap (Figure~\ref*{fig:5}, above, left panel), which lowers the linear
correlation without disturbing the ordering. (Though this is a reading
of their published figure rather than a measured comparison.) Arousal
is, on this evidence, the weaker and noisier of the two axes in both
models' representations of Claude-rendered emotion.

\subsection{Labeled circumplex
placements}\label{labeled-circumplex-placements}

The axis structure of \S5.1 does not by itself guarantee that individual
named emotions land where affect psychology would place them, so we test
placement directly. When we plot the 171 vectors on the first two
components and label a representative set of emotions across the space,
Gemma's constellation reproduces the affective circumplex reported in
Figure 57 of \citet{sofroniew2026emotion}. High-arousal negative states
(outraged, annoyed, nervous) sit upper-left and high-arousal positive
states (playful, excited, cheerful) sit upper-right. Warm low-arousal
positive states (loving, compassionate, hopeful) sit lower-right and
low-mood low-arousal states (depressed, tormented, sullen) sit
lower-left.

\subsection{The valence balance of the vocabulary is a word-list
property}\label{the-valence-balance-of-the-vocabulary-is-a-word-list-property}

93 of the 171 emotions in the circumplex (Figure~\ref*{fig:6}) fall on the negative
half of Gemma's valence axis, though the distribution's shape leans the
other way. (The positive emotions are fewer but reach further out, a
slight positive skew.) \citet{sofroniew2026emotion} make no claim about
the valence balance of their word list, but the circumplex count invites
a question: how much of this negative tilt comes from the inherited 171
emotions and how much comes from Gemma's geometry itself?

\begin{center}
\begin{minipage}{\linewidth}
\centering
\includegraphics[width=\linewidth,trim=0 0 0 38,clip]{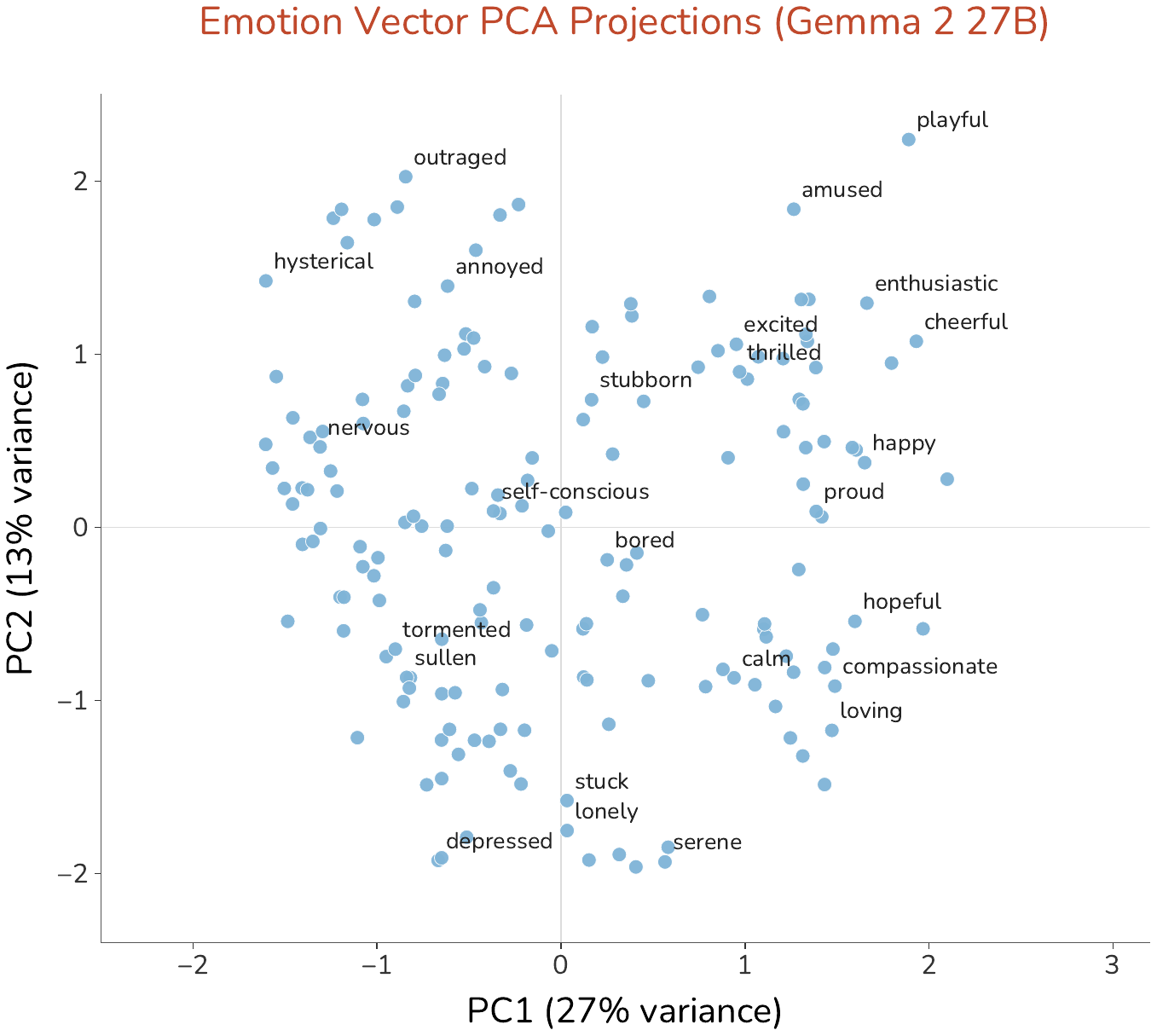}
\captionof{figure}{\textbf{The affective circumplex in a base model's pretrained representations.} 171 emotion vectors on the L31 principal components (unsupervised PCA). Cf. \citet{sofroniew2026emotion}, Figure 57.}
\label{fig:6}
\end{minipage}
\end{center}

A residual test offers an answer. We regress each emotion's PC1 position
on an independent human valence rating and ask whether Gemma places
emotions further toward the negative side than their ratings predict.
The 45 overlapping emotions in \citet{russell1977evidence} are too few
to power the test, but NRC-VAD v2.1 covers 161 emotions that overlap
with the 171 in \citet{sofroniew2026emotion} (see also \S9.3). Each
emotion's gap between where Gemma places it and where NRC-VAD ratings
predict shows no negative lean: the median residual is +0.05 standard
deviations, toward positive; the residual skew is +0.20; and a Wilcoxon
signed-rank test does not reject a zero median (p = 0.99). Valence
alone, in fact, predicts more emotions on the negative side (101 of 161)
than actually land there in Gemma's circumplex (88 of 161).

The negative tilt is therefore a property of the inherited word list,
not of Gemma's geometry. The residual test was specified in advance,
late in drafting: we report the one specification we ran, without
searching over lexicons or alternative regressions.

\subsection{Vector norms track
valence}\label{vector-norms-track-valence}

We extend beyond \citet{sofroniew2026emotion} to ask whether the L2 norm
(Euclidean length) of each emotion's vector carries affective
information. Across 171 emotions, each vector's L2 norm correlates with
the valence component at r = +0.41 (+0.37 on the raw vectors). It is
null on signed arousal (r = +0.15, not significant) and tracks arousal
\emph{extremity} weakly (absolute second component, r = +0.24). The
valence coupling is real but moderate because the norm extremes are
valence-mixed: the lowest-norm emotions are uniformly diffuse negatives
(troubled, upset, distressed, bitter), while the highest-norm set mixes
high-positive (playful, kind) with high-negative (hysterical) states. We
report this at the population level only and read norm as
representational scale or readout gain, not as emotional importance or
intensity.

The norm's behavior across depth also speaks to the choice of analysis
layer. Mean norm grows almost monotonically and peaks at the final
layer, so L31 is not the maximum-amplitude layer. (Recall that L31 was
fixed at $\sfrac{2}{3}$ depth by replication design.) The coefficient of variation of
norms is flattest across the late-middle band, one of the stable-band
signals collected in Table~\ref*{tab:2} (\S6.5).

\subsection{Natural-text co-activation preserves the
geometry}\label{natural-text-co-activation-preserves-the-geometry}

\citet{sofroniew2026emotion} assemble their held-out corpus for
activation demonstrations (e.g., their Figure 1), but do not compare
their geometry against it. As a check that Gemma's emotion geometry is
not simply an artifact of the Claude-rendered stimulus set, we ask
whether the vectors' pairwise cosine structure (\S4.1) reappears in how
they co-activate on external human text.

The token pool for this test is limited by what the activation sweep
stores. While the sweep (\S8) covers the full 47,274-document held-out
corpus (\S3.6), it records complete per-token projections only for each
vector's top-activating documents and keeps summary statistics for
everything else. The top-activating documents (723 of them, totaling
831,703 tokens) remain available at token resolution, so we run the
analysis on that pool. (A full-corpus comparison would require
re-running the sweep with per-token storage throughout, and we record it
as an open extension in \S10.2.)

For every pair of emotions, we correlate the two vectors' token-level
activation profiles across the pool, giving a co-activation matrix over
all 14,535 pairs, and we compare it entry by entry with the L31
cosine-similarity matrix of the vectors. This is a representational
similarity analysis (RSA) in the sense of \citet{kriegeskorte2008rsa}.
These two matrices agree robustly at r = 0.907 (Spearman 0.907).

Two checks bolster the agreement further. First, centering within each
document, which removes document-topic covariance, costs only 0.012 (r =
0.895). Second, excluding every token at which any emotion exceeds its
own 99th percentile, which removes the shared extreme tail, costs little
(r = 0.894). The remaining disagreement is itself structured rather than
noise, as it concentrates among low-arousal negative near-synonyms.
These emotions sit close in the geometry but tend to occur one at a time
in running text, so their co-activation understates their geometric
similarity.

This extension complements the human-norm comparison of \S5.2 on a
different channel. That comparison ties the model's two leading axes to
human affect ratings, while this one shows the full pairwise structure
governing token-level behavior outside the extraction corpus. On the
top-activating documents within the held-out corpus, Gemma's vector
geometry is expressed in behavior on natural text.

\section{Layer dynamics}\label{layer-dynamics}

\citet{sofroniew2026emotion} track emotion geometry across layers. We
replicate their fourteen-layer stability analysis (\S6.1), then extend it
with a finer sweep of all 46 layers and an account of its final-layer
transition (\S6.2). Gemma's localization is sharp in onset and broad in
extent: a late-middle band whose geometry, neighborhoods, and affect
axes, collected in Table~\ref*{tab:2}, are all but interchangeable from layer to
layer. The remaining analyses extend across depth: we track variance
allocation (\S6.3) and affect-axis consolidation (\S6.4), corroborate the
layer choice with neighborhood stability (\S6.5), and close with an
embedding-layer baseline isolating what depth builds (\S6.6).

\subsection{Cross-layer representational
stability}\label{cross-layer-representational-stability}

We measure how the emotion geometry changes across depth by
representational similarity (RSA): at each layer we compute the pairwise
cosine matrix of the 171 vectors, and we compare each layer's matrix to
L31's. The comparison is thus a cosine similarity of the two
cosine-similarity matrices, following \citet{sofroniew2026emotion}.
Figure~\ref*{fig:7} is the replication counterpart matching the original's
sampling, with all pairwise comparisons among fourteen evenly spaced
layers. We use Gemma layers 12--40.

\begin{center}
\begin{minipage}{\linewidth}
\centering
\includegraphics[width=\linewidth,trim=0 0 0 24,clip]{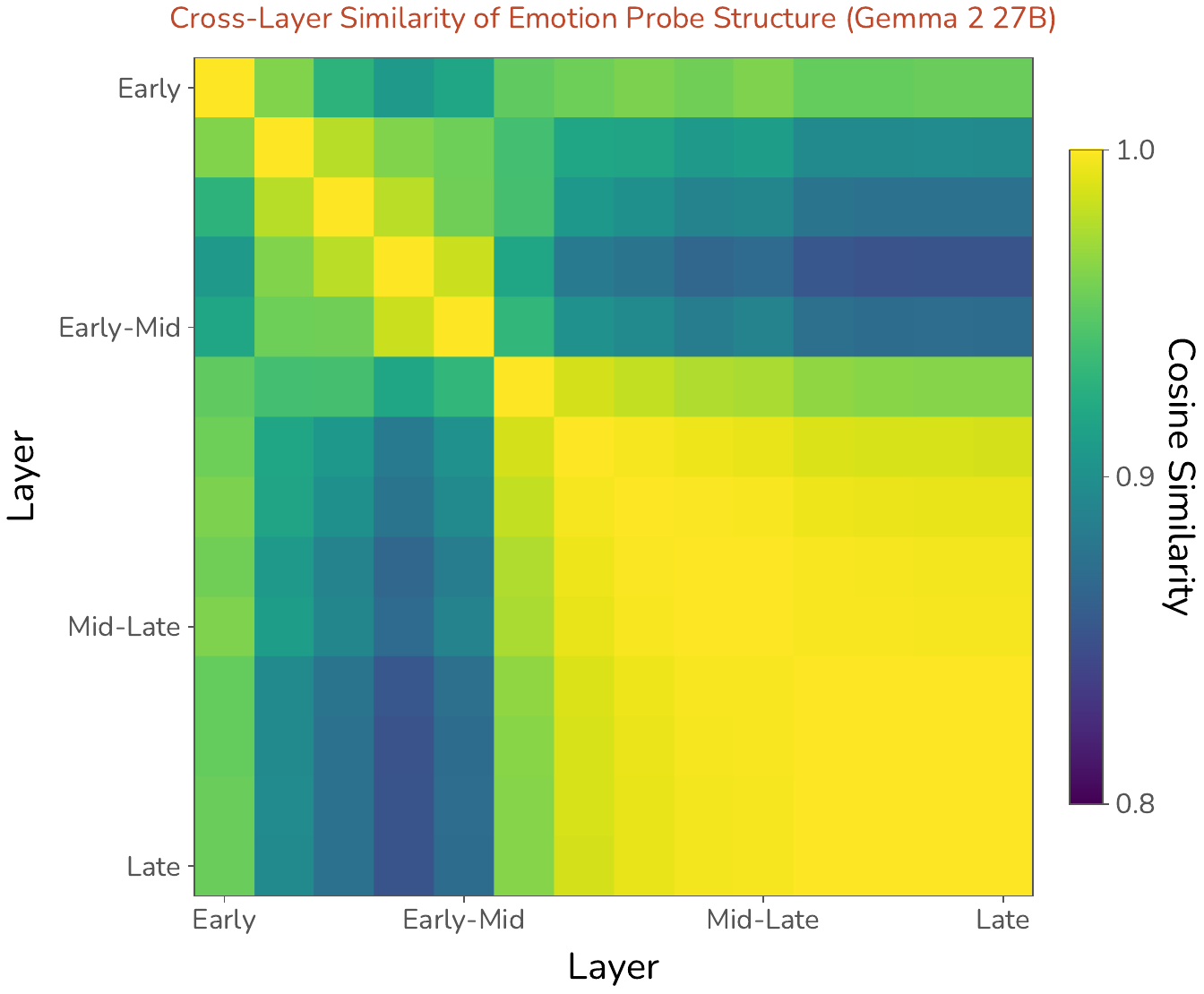}
\captionof{figure}{\textbf{Cross-layer representational similarity of the emotion geometry.} Pairwise cosine-similarity matrices of the 171 vectors at fourteen evenly spaced Gemma layers (12–40), compared across layers by the cosine similarity of those matrices. Cf. \citet{sofroniew2026emotion}, Figure 9.}
\label{fig:7}
\end{minipage}
\end{center}

At this resolution Claude and Gemma agree on the broad reading. They
both present relatively stable geometry through most of the model,
particularly from the early-middle layers onward, with the $\sfrac{2}{3}$-depth layer
inside the stable region. The visible difference at this sampling is
that Gemma's emotion geometry has an early regime and a late one, where
Claude's fades smoothly from its earliest layers into a single plateau
(\citealt{sofroniew2026emotion}, Figure 9). This, though, is an
observation of published figures rather than a measured comparison.

\subsection{The 46-layer trajectory and the final-layer
transition}\label{the-46-layer-trajectory-and-the-final-layer-transition}

As an extension, we also sweep all 46 of Gemma's layers, comparing each
layer's matrix to the canonical L31's. This allows us to understand the
boundary between Gemma's early and late regimes and capture a finer
reading of the geometry's trajectory across depth.

Similarity to L31 remains moderate through the early and middle layers
at approximately 0.89 across L0--11 and 0.90 across L12--21. It rises
sharply across L22--26, reaching 0.98, though not discontinuously: the
sharpest adjacent step, L21$\rightarrow$L22, still registers 0.94. It then
approaches identity across the canonical late-middle band, reaching
0.999 across L27--34 and remaining at 0.998 through L35--40. Similarity
falls to 0.949 across the final readout layers, L41--45. (As
confirmation that the choice of similarity measure is not doing the work
here, scoring the matrix comparison by Pearson correlation instead of
cosine similarity changes each value by at most about 4$\times$10$^{-}$$^{5}$.)

The closing dip has a mechanical source, revealed through the norm of
the residual-stream activation, averaged over the extraction tokens at
each layer. Gemma's mean residual-stream norm grows strongly and almost
monotonically with depth, from roughly 940 at L0 to 3,300 at L12, 19,400
at L31, and 42,900 at the final layer. This monotonic growth is generic
transformer behavior. The informative feature is a single non-monotonic
event near the end: the norm drops by about 4,100 from L42 to L43
(29,200 $\rightarrow$ 25,100), the trajectory's only large downward step, a property
of Gemma's final normalization (the same large gains that matter for the
logit lens, \S7). The alignment between the two analyses is close but not
exact: the similarity dip begins at L41, and the norm step sits at
L42$\rightarrow$L43, one layer into the dip. We treat the geometry at L42 onward as
non-comparable to L31, since projection ratios diverge in the final
normalization layers.

The boundaries between the trajectory's three parts depend on the
metric, as the difference between the similarity dip and the norm step
illustrates. For this reason, we report each boundary with its
corresponding measure. The 46-layer trajectory also resolves structure
at a finer grain than the fourteen-layer sweep of
\citet{sofroniew2026emotion}, mentioned previously, making it a
characterization of Gemma with no point of comparison rather than an
observable cross-model divergence.

\subsection{Variance allocation across
depth}\label{variance-allocation-across-depth}

Given that geometry survives across layers, we ask whether the geometric
variance explained by the leading component is constant across depth. To
answer, we fit a fresh PCA over the 171 cleaned vectors at each of the
46 layers and record the share of variance the leading component carries
at each one. The share of variance follows an asymmetric U-shape: it
peaks early, at 43.0\% at L10, the maximum across all 46 layers, and
remains elevated through the mid band. It then falls to its lowest point
through the canonical band (a mean of approximately 27\% across L27--34,
with L31 at 26.732\%) before climbing again toward the final layers,
about 42\% at the last layer itself, without regaining its early height.
The L41--45 band mean stays lower still, at 32\%, because the rise
concentrates in the last few layers. Table~\ref*{tab:2} (\S6.5) shows share of
variance averages across the 46 layers in six intervals.

The canonical band (L27--34) is least dominated by valence, and it is
where the geometric space looks most like a two-axis circumplex. This
band also includes the one comparable point we have with
\citet{sofroniew2026emotion}, where the valence explains
\textasciitilde27\% of the variance at $\sfrac{2}{3}$-depth of Claude's layers, while
it explains 26.7\% of Gemma's at L31 (\S5.1).

\subsection{When the affective axes
consolidate}\label{when-the-affective-axes-consolidate}

\S5.1 and \S5.2 establish the circumplex at the $\sfrac{2}{3}$-depth layer, but a
single layer cannot show \emph{when} that structure forms. If valence
and arousal lead at L31, do they lead at every depth, or does the
circumplex assemble somewhere along the way?

At each layer we fit the PCA over all 171 vectors, restrict to the 45
Russell--Mehrabian overlap emotions, and correlate the leading
components with human pleasure and arousal. This is the same global
fit-then-restrict pathway \citet{sofroniew2026emotion} use, applied
layer by layer. The pathway matters: fitting a separate PCA on the
45-emotion slice at each layer, rather than restricting the full-set
PCA, misplaces arousal. (This is the layerwise counterpart of the scale
condition in \S5.2.)

Valence rises across depth from about 0.45--0.55 early to a high plateau
across L27--43, with a canonical-band mean of 0.735, a peak of 0.749 at
L34, and 0.720 at L31. It is the stable, monotonic axis through this
interpretable range, though it falls off in the final layers along with
everything else read from that terminal region (\S6.2). Arousal, however,
follows a different path. Held to one fixed sign convention across
layers, arousal's correlation with PC2 flips between positive and
negative before the seam. It is negative at L10 ($-$0.28), turns
significantly positive in the mid-teens (L16 +0.44, L17 +0.47), and
returns to negative at L25 ($-$0.38) and L26 ($-$0.41), with bootstrap
intervals excluding zero at each of these (only marginally at L10), and
indistinguishable from zero at several other early layers. Arousal turns
stably positive from L27 onward, reaching 0.67 at L31 (canonical-band
mean 0.646) and climbing to a peak of 0.74 at L38. L31 falls inside this
resolved regime, which is why the circumplex reads cleanly there. (The
positive early and mid arousal entries in Table~\ref*{tab:2} are band-level
correlations reported without this fixed sign. Holding one sign across
all layers, as we do here, surfaces the early flips that are concealed
by band-level aggregates.)

Arousal's instability before the seam is not just a matter of sign. It
is distributed across the PC2/PC3 plane rather than consistently
aligning with the second component alone. Through the first twenty-seven
layers (L0--26), its best-aligned component is PC1 or PC3 rather than
PC2. (At L26 it sits mostly on the third component.) Even once it
resolves into the PC2/PC3 plane, the best combination of those two
components captures it better than PC2 alone: multiple correlation R =
0.71 at L26, 0.75 at L31, and 0.81 at L40, against 0.41, 0.67, and 0.73
for PC2 alone. Adding PC1 contributes almost nothing. A variance control
confirms this late resolution is developmental rather than an artifact
of a weak component. The second component's own variance share stays
flat across depth while its correlation with arousal swings.

This resembles behavior \citet{sofroniew2026emotion} describe in Claude
but do not quantify. (They write that arousal occupies ``a mix of the
second and third PCs, depending on the layer.'') Our sweep finds the
same layer-dependent PC2/PC3 mixing in Gemma, so it is a point of
cross-model convergence. Together with the scale-dependence of \S5.2,
this depth profile explains why arousal fails the stability criterion of
\S1. The correlation is real and strengthens with depth, but it does not
hold as a stable feature across emotion-set sizes and depth regimes. So
we do not classify arousal as replicated.

\subsection{Neighborhood stability and the convergent case for
L31}\label{neighborhood-stability-and-the-convergent-case-for-l31}

\S6.2, \S6.3, and \S6.4 are three different population summaries
corroborating that the pre-committed L31 sits in a representative part
of Gemma's emotion geometry. A summary can stay stable in aggregate,
though, while individual emotions drift underneath it. This section
checks individual emotions directly, asking whether each emotion's own
neighborhood remains stable across depth.

First, we track emotion neighbors with a retention score: the Jaccard
overlap between each emotion's top-5 neighbors at a given layer and its
top-5 neighbors at L31, averaged across all 171 emotions. That score
sits at 0.56--0.59 through the early and middle layers, rises to 0.79
across the seam, and reaches 0.95 through the canonical band. It holds
at 0.90 through the late band before a partial churn back down to 0.77
at the very end. In short, top-5 cosine-neighbor sets churn early and
then stabilize sharply.

Second, at the per-emotion level, we measure the overlap between an
emotion's top-10 neighbors at a given layer and its top-10 neighbors at
L31, out of a possible 10. That overlap reaches 9.76 of 10 (median 10)
across the canonical band L27--34, essentially identical to L31
throughout. It falls to 8.82 at the seam and to about 7 early and mid.
An emotion's neighbor list only partly matches L31's early on, then
becomes almost indistinguishable from it through the canonical band.

Analysis of individual emotions provides additional evidence that L31 is
representative of the canonical band, though within that L27--34 band no
single layer is privileged over its neighbors. The least stable emotions
under this measure are ambiguous boundary cases (bored, infatuated,
alert, restless, patient), with infatuated also on \S4.3's seed-stability
list. The most stable are prototype emotions at cluster cores. The split
is interpretable, not evidence of a competing taxonomy.

The per-emotion view also clarifies the drop in Gemma's final layer.
Neighbor sets remain intact into L43 even where residual norms have
fallen. An emotion's nearest neighbors at L31 are typically still its
nearest neighbors at L43, so the late change is a magnitude and
output-space event (\S6.2), not a semantic reorganization. Neighbor
identity is, on this evidence, more stable than either global geometry
or norm magnitude.

Table~\ref*{tab:2} collects the band means of the depth analysis: representational
similarity to L31 (\S6.2), neighbor stability by both measures above,
leading-component variance (\S6.3), affect-axis alignment on the
global-PCA pathway (\S6.4), and norm variability (\S5.5). Overlap means
include L31's own layer. Because arousal's sign is unstable before the
seam under a fixed global convention (\S6.4), the pre-seam arousal cells
report magnitudes rather than signed correlations. Adjacent bands blend
into one another rather than break cleanly: early and mid differ by only
0.024 on Jaccard retention and 0.011 on RSA, and the trajectories are
non-monotonic, with the mid band holding the highest PC1-variance band
mean. The RSA and Jaccard columns restate the locked per-band values
reported above; the remaining cells come from the released per-layer
artifacts via a released derivation script.

\begin{center}
\begin{minipage}{\linewidth}
\small
\captionof{table}{\textbf{Band means of the depth analysis.} Seven signals over the six reporting bands, cleaned vectors throughout (signals, conventions, and provenance in \S6.5). Band means summarize continuous per-layer trajectories; the boundaries are a reporting convention, not sharp transitions. The canonical band containing the pre-committed L31 is where every signal sits at or near its optimum simultaneously.}
\label{tab:2}
\begin{tabularx}{\linewidth}{@{}>{\hsize=1.780\hsize\raggedright\arraybackslash\hyphenpenalty=10000\exhyphenpenalty=10000}X>{\hsize=0.889\hsize\centering\arraybackslash}X>{\hsize=0.889\hsize\centering\arraybackslash}X>{\hsize=0.889\hsize\centering\arraybackslash}X>{\hsize=0.889\hsize\centering\arraybackslash}X>{\hsize=0.889\hsize\centering\arraybackslash}X>{\hsize=0.889\hsize\centering\arraybackslash}X>{\hsize=0.889\hsize\centering\arraybackslash}X@{}}
\toprule
\textbf{Band} & \textbf{RSA→L31} & \textbf{Top-5 Jaccard} & \textbf{Top-10 overlap} & \textbf{PC1 var \%} & \textbf{Valence r} & \textbf{Arousal r} & \textbf{Norm CV} \\
\midrule
early L0–11 & 0.887 & 0.56 & 6.93 & 29.11 & 0.465 & 0.229 & 0.29 \\
mid L12–21 & 0.898 & 0.59 & 7.09 & 35.19 & 0.476 & 0.245 & 0.29 \\
seam L22–26 & 0.978 & 0.79 & 8.82 & 29.05 & 0.685 & 0.272 & 0.24 \\
canonical L27–34 & 0.999 & 0.95 & 9.76 & 27.06 & 0.735 & 0.646 & 0.22 \\
late L35–40 & 0.998 & 0.90 & 9.55 & 27.46 & 0.735 & 0.729 & 0.22 \\
final L41–45 & 0.949 & 0.77 & 8.47 & 31.88 & 0.648 & 0.646 & 0.26 \\
\bottomrule
\end{tabularx}
\end{minipage}
\end{center}

\subsection{The embedding-layer
baseline}\label{the-embedding-layer-baseline}

So far, \S6 has tracked how Gemma's geometry changes as computation
proceeds through its 46 layers. What does the geometry look like before
any layer runs at all? We answer this with a bag-of-embeddings baseline.
For each emotion, we look up Gemma's static input embeddings for the
tokens in the same 205,200 stories and 500 neutral dialogues. We then
apply the paper's pooling rule, taking the mean from the 50th token
onward.

Vector construction and confound cleanup use the same
difference-of-means and cleanup functions applied at L31. The cleanup
basis, though, is computed fresh from the neutral dialogues' own
embeddings rather than carried over from L31. This way the comparison
isolates depth, not a different pipeline. We report the raw,
assumption-free result as primary and the cleaned variant alongside it
(Table~\ref*{tab:3}). The embedding table itself is read directly from the weights
of \texttt{google/gemma-2-27b}, without loading the full model. This is
an extension beyond \citet{sofroniew2026emotion}.

\begin{center}
\begin{minipage}{\linewidth}
\small
\captionof{table}{\textbf{The embedding-layer baseline against L31}, cleaned basis unless noted. No forward pass at the embedding row; L31 values are the results reported in \S4.1, \S4.2, \S5.1, and \S5.2.}
\label{tab:3}
\begin{tabularx}{\linewidth}{@{}>{\hsize=1.647\hsize\raggedright\arraybackslash\hyphenpenalty=10000\exhyphenpenalty=10000}X>{\hsize=0.760\hsize\raggedright\arraybackslash}X>{\hsize=0.593\hsize\raggedright\arraybackslash}X@{}}
\toprule
\textbf{Signal} & \textbf{Embedding (no depth)} & \textbf{L31} \\
\midrule
Valence, PC1$\leftrightarrow$pleasure & 0.479 (raw 0.378) & 0.720 (raw 0.681) \\
Arousal, PC2$\leftrightarrow$arousal & 0.179 (raw 0.282) & 0.668 (raw 0.619) \\
PC1 / PC2 variance & 23.1\% / 14.5\% & 26.7\% / 13.4\% \\
RSA to L31 & 0.882 & 1.000 \\
k=10 families, ARI vs. locked L31 partition & 0.481 & cross-seed self-ARI 0.686 (min 0.467) \\
Neighbor cosine, afraid→scared & 0.960 & 0.987 \\
Neighbor cosine, joyful→happy & 0.916 & 0.977 \\
Neighbor cosine, compassionate→empathetic & 0.954 & 0.965 \\
\bottomrule
\end{tabularx}
\end{minipage}
\end{center}

As Table~\ref*{tab:3} demonstrates, several parts of the L31 geometry are already
present in Gemma's static embeddings. Afraid and scared have a cosine
similarity of 0.960 before any transformer layer runs. The
embedding-level family partition resembles the locked L31 partition
about as closely as L31's least-agreeing seeds resemble each other (ARI
0.481, against a cross-seed mean of 0.686 and a minimum of 0.467). The
embedding baseline's RSA to L31 is 0.882, close to the L0--11 mean of
0.887 reported in Table~\ref*{tab:2}. By this measure, the first twenty-odd layers
preserve a geometric structure already present in the embedding table.
The leading-component variance split and much of the valence structure
are also present at the embedding layer.

Arousal, again, is different. Its correlation at the embedding layer is
only 0.179, compared with 0.668 at L31. Among the signals measured here,
it is the only one that is largely absent from the embedding baseline
and strong at the analysis layer. In other words, the network has to
build arousal for itself.

\section{Vocabulary readout}\label{vocabulary-readout}

Having characterized the geometry (\S\S4--6), we now ask what the emotion
vectors yield at the vocabulary interface. Which tokens does each vector
promote when projected to the model's output space, and at what depth
does that readout become legible? This section reports the
direct-unembedding table that parallels the original study's Table 1
(\S7.1), a methods finding about handling the projection on raw Gemma
weights (\S7.2), and the depth at which the readout switches on (\S7.3).

\subsection{Direct-unembedding
readout}\label{direct-unembedding-readout}

Table 1 in \citet{sofroniew2026emotion} projects twelve emotion vectors
through the unembedding matrix and reports the top five promoted tokens
and bottom five suppressed tokens for each. Their table's central
finding is that each vector promotes tokens semantically related to its
emotion (e.g., Guilty's top five promoted tokens are guilt, conscience,
guilty, shame, blamed). \citet{sofroniew2026emotion} describe the
operation as measuring ``the direct effects of each emotion vector on
the model's output logits through the unembed.'' We read this to mean
the direct projection $v \cdot W_U$ (i.e., the vector against the
unembedding matrix, with no other machinery) and apply it to Gemma's
L31. Table~\ref*{tab:4} is the result.

\begin{table}[tbp]
\small
\caption{\textbf{Direct-unembedding readout at L31 (direct projection $v \cdot W_U$).} Tokens verbatim (leading word-boundary markers dropped). Cf. \citet{sofroniew2026emotion}, Table 1.}
\label{tab:4}
\begin{tabularx}{\linewidth}{@{}>{\hsize=0.561\hsize\raggedright\arraybackslash\hyphenpenalty=10000\exhyphenpenalty=10000}X>{\hsize=1.147\hsize\raggedright\arraybackslash}X>{\hsize=1.292\hsize\raggedright\arraybackslash}X@{}}
\toprule
\textbf{Emotion} & \textbf{Promoted (↑, top-5)} & \textbf{Suppressed (↓, bottom-5)} \\
\midrule
\textbf{Happy} & excited, excited, excit, joy, happiness & gentil, loyal, {\FBdev क्र}, Elegant, politely \\
\textbf{Inspired} & neglected, excited, inspired, dusty, ideas & {\FBserif демон}, esfuerzo, Liar, ulaski, {\FBtib ང་} \\
\textbf{Loving} & goofy, tenderness, {\FBcjk 怱}, corações, licate & politely, Einstellungen, mantiene, aumentado, Gdzie \\
\textbf{Proud} & proud, proudly, prou, proudest, beaming & fijn, {\FBcjk なのに}, kecelakaan, fijne, alleviate \\
\textbf{Calm} & curious, satisfied, mindful, enjoy, entstand & berusaha, terus, {\FBcjk みました}, {\FBsym ❢}, sentía \\
\textbf{Desperate} & desperate, desesper, desperation, {\FBcjk 優しい}, afford & ǒ, appartiennent, {\FBcjk ￣}), lamentable, inconven \\
\textbf{Angry} & anger, Anger, Anger, rage, angrily & smiles, maha, {\FBheb ָּ}, {\FBcjk くて}, {\FBmathf 𝑰} \\
\textbf{Guilty} & guilt, carteles, irée, {\FBcjk 졍}, etl & surprisingly, seriously, dramatically, dramatic, former \\
\textbf{Sad} & loneliness, depression, sadness, Lon, Depression & {\FBcjk 犹如}, {\FBarab يكون}, ICAL, {\FBori\XeTeXglyph 2}
, entially \\
\textbf{Afraid} & panic, panic, paranoia, Panic, Panic & exasperated, frustrated, eyebrow, baffled, understood \\
\textbf{Nervous} & Anxiety, nervousness, nervously, Anxiety, anxiety & smiles, isier, olhos, exhausted, anyan \\
\textbf{Surprised} & shock, \emojipng{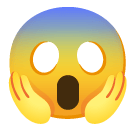}, shock, {\FBtib ང་}, \emojipng{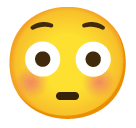} & nanti, afterwards, tomorrow, smiles, insomnia \\
\bottomrule
\end{tabularx}
\end{table}

The finding broadly replicates on Gemma: the top tokens for Nervous are
recognizably Anxiety, nervousness, nervously, Anxiety, anxiety, and the
top five for Proud are proud, proudly, prou, proudest, beaming. As
``prou'' illustrates, relevant word fragments appear among the hits, as
do emojis, non-English words and fragments, and tokens with no evident
connection to the emotion. For example, the top tokens for Surprised
include \emojipng{emoji_scream} and {\FBtib ང་} (the Tibetan letter nga plus a syllable mark), while
Guilty's include irée (a French word-ending, as in soirée and désirée)
and {\FBcjk 졍} (an uncommon Korean syllable).

Notice that a tidy list of complete words in English would be a more
surprising result than the mixed output we see in Table~\ref*{tab:4}. The emotion
vectors are extracted over stories in English, but each emotion
concept's nearest tokens are drawn from Gemma's whole vocabulary of
multilingual words and fragments, emoji, mathematical symbols, and
stylized characters. We report the readout qualitatively, therefore, and
do not interpret the suppressed columns row by row, as they are more
heterogeneous. \citet{sofroniew2026emotion} do the same.

Still, Gemma's readout is noisier and more multilingual than Claude's,
though Claude's includes fragments and non-English tokens, too. For
example, Claude's top five promoted tokens for Loving are treas, loved,
{\FBcjk ♥}, treasure, and loving, while Gemma's are goofy, tenderness, {\FBcjk 怱} (a rare
Chinese character), corações (Portuguese for ``hearts''), and licate
(English fragment appearing in delicate or complicate). Gemma's most
on-theme rows are the negative, high-arousal emotions (anger, fear,
anxiety, sadness, desperation). Its positive-affiliative rows (inspired,
loving, calm) are the least uniformly on-theme.

\subsection{Applying Gemma's final RMSNorm to the vectors before
projecting}\label{applying-gemmas-final-rmsnorm-to-the-vectors-before-projecting}

\citet{sofroniew2026emotion} did not report whether they applied
normalization to the emotion vectors before projection, so our Table~\ref*{tab:4}
applies none. To test the alternative, however, we apply Gemma's final
RMSNorm to the vectors before projecting at L31. Angry remains clean and
Calm arguably improves, but guilt drops out of Guilty's top five. Sad's
promoted tokens fill with fragments, and Afraid and Surprised both open
on fragments with no evident relation to either emotion. In short, the
readout's quality decreases substantially. (We do not reproduce the
table here, but the released lens ingredients and scripts regenerate it,
along with other variants we examined. See \S11.)

The normalized version exposes a mismatch between the instrument and the
object. Gemma's final-RMSNorm weight is large by design, with
per-dimension gains up to roughly 82$\times$. Applied to a genuine
residual-stream state, those gains are stabilized by the state's
baseline variance, but a difference-of-means vector has no such
baseline, so the gains amplify high-gain directions unrelated to the
emotion. The normalized lens is the right instrument for genuine
residual states, but for reading a synthetic vector's alignment with the
token directions, the direct projection is the right instrument.

Two notes for replicators. First, what ``the plain projection'' denotes
depends on how the weights are loaded: in TransformerLens's default
processed regime the final norm is folded into $W_U$, so
$v \cdot W_U$ would silently include normalization. In the
raw-weights regime we use (\S3.1), it does not, and Gemma-2's norm must
be applied explicitly in its (1~+~w) form when a normalized lens is
wanted. (Both the unfolded export and the (1~+~w) form are attested by
the export's provenance metadata and the lens code.) Second, a
diagnostic trap: a near-zero mean normalization weight usually signals
the (1~+~w) storage form, but the final norm's weights are large by
construction. Confirm the storage form from the architecture, not from a
weight-mean test.

\subsection{When the readout becomes
legible}\label{when-the-readout-becomes-legible}

Table~\ref*{tab:4} asks a geometric question of the readout. Moving from L31 to a
later layer on the direct projection, for instance, produces visibly
cleaner rows, with Loving opening on tenderness and Sad on sadness. We
ask, then, two questions about depth. At what layer does the readout
become legible? And is its arrival related to the seam in Gemma's
geometry at L22--26 (\S6.2)?

This is an extension from the original study, so with no inherited
legibility values we define the measure: the median rank of each
emotion's own word among the 256,000 tokens its vector scores, computed
at every layer with the normalized lens. While the norm distorted the
geometric question, omitting it here would skip a real step between
residual stream and logits. In the readout, ``unreadable'' describes
median ranks around 20,000--24,000 (the emotion's own word buried below
the top 20,000 tokens), while ``legible'' describes ranks in the low
thousands. ``Onset'' is the defined event: the first layer where the
median crosses below 10,000, or roughly the top 4\% of the vocabulary.

We find that the emotion vectors are effectively unreadable through
L0--21 and become lexically legible across the same region the geometry
consolidates. Median own-word self-rank sits at roughly 20,200 at L21
and 23,700 at L22, then collapses to roughly 5,400 at L23 and 3,900 at
L24, with onset, defined as the first layer at which the median
self-rank falls below 10,000, at L23. The onset at L23 lands one layer
past the sharpest step in the cross-layer geometry, L21$\rightarrow$L22 (\S6.2).
Representational geometry and output readability locate the seam within
a single layer of each other.

Early illegibility alone would not be surprising. Logit-lens readouts
degrade away from the final-layer basis \citep{belrose2023tuned}, and
other studies report mid-depth onsets of decodability
\citep{wendler2024llamas,lad2024stages}. The important result here is
that readability begins in the same layer range where a lens-independent
analysis finds the geometry reorganizing.

To test whether the onset is structural or lexical, we regress each
emotion's onset layer on its word's frequency (Zipf value). This
explains only a small minority of the variance (R$^{2}$ $\approx$ 0.02 at a
self-rank-below-1000 threshold, $\approx$ 0.07 at a stricter below-100
threshold). Decisively, even the most frequent third of the emotion
words do not become readable early, onsetting at the seam like the rest.
Emotion readability therefore switches on where the geometry
consolidates, not where common words sit in the vocabulary. This result
holds at the population level only. A per-family staging effect failed a
construct-validity re-test under the frozen specification. The apparent
staging traced back to the family definition itself, and we record it as
a method gap rather than a finding (\S10).

\section{Activation on natural
text}\label{activation-on-natural-text}

This section asks what the emotion vectors do on naturalistic text, and
it is motivated by a warning from \citet{sofroniew2026emotion} that a
vector might track incidental details of the elicitation setting rather
than the emotion concept. We investigate that warning directly: \S8.1
asks where in held-out documents the vectors activate most strongly and
quantifies a token-level confound in the answer. \S8.2 asks where those
activations come from in the corpus. \S8.3 asks whether every vector
carries signal above the noise of ordinary text and collects what
survives the confound. All results in this section come from the
held-out evaluation corpus of \S3.6 (47,274 documents, 25.1M swept tokens
at L31, unit-normalized projections) and are reproducible from the
released scripts and data (\S11).

\subsection{What the vectors maximally activate
on}\label{what-the-vectors-maximally-activate-on}

Our Figure~\ref*{fig:1} (\S1) replicates Figure~\ref*{fig:1} in \citet{sofroniew2026emotion},
which demonstrates per-token activation shading on held-out text floored
at each vector's own 90th percentile. Why a given stretch of text would
land among these highest-token activation windows, however, is not
always immediately apparent. While selecting printable windows for
Figure~\ref*{fig:1}, for instance, we found that most of the highest-activating
passages for desperate and surprised do not contain relevant emotion
content at all. Instead the activations land on negation contractions
(don't, didn't, couldn't) and on citation or figure tokens. (A
combustion engine patent appeared among the highest-activating passages
for both desperate \emph{and} surprised.) The method that informs Figure
1, then, invites a more pointed extension: when a vector's activation
peaks, what token does it actually land on?

This is an extension, so we fix and release the classification here. A
peak token counts as structurally non-conceptual if it is a bare
numeral, punctuation or a symbol, a function word, or whitespace. This
is a deliberate floor rather than a complete confound count. We do not
hand-mark the remaining content-word peaks as on- or off-theme, because
judging a bare peak token requires supplying context the token does not
carry. (Is what's \emph{pounding} a heart or a hammer?) A marked total
would measure the reader's inference rather than the vector. The true
confound rate is therefore higher than what we report and remains
unquantified.

At rank 1, 52.6\% of peak tokens (90 of 171) are structurally
non-conceptual: a bare numeral (21.1\%), punctuation or a symbol
(13.5\%), whitespace (11.7\%), or a function word (6.4\%). The rate
remains high below rank 1: across all 3,420 top-20 windows, 48.4\% of
peaks are structurally non-conceptual. A random-direction control sets
the baseline. When 200 random directions are run through the same
decode-and-classify pipeline, they peak on structurally non-conceptual
tokens 66.0\% of the time, against 51.5\% for the emotion vectors. The
structural confound is therefore partly generic to extreme projections
on this corpus rather than specific to emotion vectors. But the emotion
vectors sit well below that random rate, so they land on conceptual,
content-bearing tokens well above chance.

The stored windows reveal why bare numerals are the largest single
category (36 of 171 emotions). The peak digit is the numeral inside an
LMSYS anonymization token, standing exactly where a person is named
(e.g., ``NAME\_1 was overjoyed''). The vector fires on an emotionally
charged span while its single strongest position lands on the
scrub-token digit. The 36 emotions where this occurs lean positive
overall, but what unifies them is social rather than valence: alongside
the high-arousal positive cluster, the set also includes negative,
other-directed emotions like embarrassed, humiliated, and outraged, plus
gratitude. All 36 turn on how one person is seen by or relates to
another, which is exactly where a conversation might explicitly name the
people involved.

This classification identifies \emph{where} the peaks land but cannot
determine \emph{why} they land there. \citet{tigges2023sentiment} find
that sentiment in transformers is partly summarized at syntactically
uninformative positions, punctuation and names among them. The
anonymized-name peaks here may simply be artifacts of the scrub-token
scheme, or they may reflect that same tendency. Under either
explanation, the peak token cannot by itself be read as the location of
the emotion concept.

The structural floor leaves 81 content-word peaks unclassified, and the
peaks collapse across related vectors on both sides of the
classification. The sharpest single collapse sits in the joy family.
Eight of its ten vectors share one identical rank-1 peak, a scrub-token
numeral in a single LMSYS document. The fear family collapses at the
window level instead. Across sixteen fear-family vectors (a fixed list
of fear terms, a subset of the 29-member Fear and Overwhelm family of
\S4.2), 320 top-20 window slots land on only 72 distinct peaks, and the
most overlapping pairs (frightened--scared, on edge--unnerved) share 19
of their 20 top windows outright. Whatever fine structure separates
these directions in the circumplex geometry (\S5) is not expressed at the
extreme tail of naturalistic activation, where the fear family behaves
as a single racing-heart detector.

That collapse could still be read charitably as semantic overlap between
similar emotions rather than a failure to track concepts at all. So we
check whether the peak lands on the emotion concept word itself when
that word is in the text. Across the top twenty highest-activating
documents for each emotion (171 x 20 = 3,420 documents), only 148
documents contain the emotion word as a whole-word string somewhere in
the document. Of those 148 documents, the single strongest token lands
on the specific emotion word only 9 times (6.1\%), roughly one time in
sixteen. Even when the emotion word is present within the text and
directly competing, the readout still passes it over.

\subsection{Where the maxima come
from}\label{where-the-maxima-come-from}

Once we identify each vector's peak-activation tokens, we can examine
the surrounding text. Do the peak tokens for 171 emotions distribute
proportionately across the four sources in the held-out corpus, or does
one source dominate beyond its share? When we compare the source of each
top-activating window against the corpus's own recorded document and
token baselines, the mix approximately tracks the corpus's own token
baseline, and no single source dominates beyond its share. We report
this qualitatively rather than as fixed percentages, because the source
mix is an extreme-value statistic of a single 25.1M-token sweep,
concentrated in the anonymization-token spikes of \S8.1 and dependent on
the particular documents that occupy the extreme tail. Consistent with
this, the vectors' mean projection on LMSYS text is if anything slightly
lower than on the other sources, so any apparent source selection at the
top of the distribution is a tail phenomenon rather than a mean-level
preference.

Per-emotion variation remains, but it too is tail-sensitive. The
emotions whose top windows come predominantly from outside LMSYS are
broad and spread across the circumplex, and the set is too
tail-sensitive to attach to individual vectors, so we report no
per-emotion source lists. The durable findings from this sweep are the
structural floor and collapse structure of \S8.1 and the magnitude
calibration of \S8.3. Where the maxima come from is, on this evidence, a
property of the corpus draw at least as much as of the emotion
directions.

\subsection{Magnitude calibration, and what survives the
confound}\label{magnitude-calibration-and-what-survives-the-confound}

If the single highest-activating token for an emotion vector is
infrequently the emotion word, we should check whether the vectors track
emotion at all. We run, then, one final check to ask whether each vector
carries in-distribution signal above the noise of ordinary text,
independent of where its peak lands. For each emotion we measure the
separation between its own-story activation and the natural-text
activation distribution, in units of the natural-text spread.

All 171 vectors separate by at least 1.01$\sigma$ (median 2.08$\sigma$; none below
1$\sigma$), so every direction carries signal above natural-text noise. Because
the vectors were extracted from these stories, strong activation on the
stories is expected. The separation from natural-text noise is the
informative quantity. It measures scale calibration on external text,
not independent semantic validity.

\citet{sofroniew2026emotion} applied a version of this check to an
alternative logistic-regression probe built to assess a character's
chronic emotional state. That probe failed both prongs of the check: its
top passages showed little discernible emotional content and its
magnitudes on natural documents were very low. The authors read the
pattern as overfitting. They also did not report the quantitative check
for its core story-based vectors. Here we do both.

On the content prong, the Gemma vectors' top passages are emotionally
apt at the region level, with the token-level peak confound of \S8.1 as
the disclosed remainder. On the magnitude prong, every vector separates
cleanly. The weakest separators are diffuse-distress emotions such as
troubled, upset, unhappy, and hurt, all below 1.2$\sigma$. They overlap with
the fear family, whose maxima often fall on shared somatic tokens
(\S8.1). Both patterns likely reflect the same cause. Generic distress
language is common in ordinary text, so these vectors separate weakly
from the natural-text baseline and collapse onto the same peaks. The
strongest separators are distinctive-register emotions (playful 3.86$\sigma$,
lazy 3.85$\sigma$, amused 3.84$\sigma$, sleepy 3.82$\sigma$).

So what survives the confound? The emotion vectors reliably find
emotional passages, but their highest-activating tokens often
misidentify where the emotional signal resides within those passages. It
is a confounded instrument at the single-token level. The confound is
systematic and measurable, and it leaves the geometric evidence for the
vectors untouched. The vectors select emotionally appropriate text,
separate cleanly from natural-text noise, and light up across whole
emotional regions even where their brightest token is an artifact.

\section{Discussion and
limitations}\label{discussion-and-limitations}

This section offers discussion and limitations of the replication study.
\S9.1 weighs what generalized from \citet{sofroniew2026emotion}, what
varied, and what the extensions add. It also grades each headline result
by the criteria set out in \S1. \S9.2 states the replication's limitations
and \S9.3 places the results within the prior and concurrent literature,
including the one apparent disagreement we take up in detail.

\subsection{Discussion}\label{discussion}

\citet{sofroniew2026emotion} include in their limitations an expectation
and a caution: ``While we expect the broad findings to generalize, the
details of our results may vary across model families, sizes, and
training procedures.'' This replication tests that expectation and
substantiates its caution. Gemma recovers the affective circumplex, the
separation of valence and arousal, the family-level taxonomy, and a
stable late-middle band containing the pre-committed layer. It differs
from the original most clearly in the cluster boundaries, and although
it recovers the arousal axis, it does so only conditionally (\S5.2,
\S6.4). The pipeline was inherited at every disclosed decision point
(\S2), so both the generalization and the variation are instructive.

Because the subject is base google/gemma-2-27b, the recovered structure
is a property of the pretrained representation of Claude-rendered
emotion (\S3.2), obtained from a model outside the Claude family. Whether
such structure survives training is a question the original could only
approach indirectly: \citet{sofroniew2026emotion} applied the
post-trained model's probes to both Claude's base and post-trained
models and found the activation patterns largely preserved (r = 0.83 on
neutral scenarios, 0.67 on challenging ones) with consistent
training-related shifts (r = 0.90). What they assumed, rather than
measured, was that the vector directions themselves stay stable.
Re-extracting the full geometry directly from a base model, albeit one
in a different family, supplies exactly the direction-level evidence
their assumption calls for.

The base-model recovery also supplies a premise concurrent work already
relies on. \citet{han2026welfare} show that a reward-derived functional
welfare axis aligns tightly with the valence axis of emotion concept
vectors extracted via Sofroniew et al.'s own story-based methodology (R$^{2}$
= 0.948). They also show that the axis predates alignment training. It
is partially present in pretrain-only models, and both RL and supervised
fine-tuning recruit it rather than create it. This presupposes exactly
the pretraining-native structure documented here.

The clearest replications are the leading component's variance share
(26.7\% versus \textasciitilde27\%), recovery of the family-level
taxonomy, and the alignment between PC1 and human valence norms (0.72
versus 0.81). The valence alignment also remains stable across
emotion-set sizes and depth. Arousal falls close to the original result
(0.67 versus 0.66 at L31), but it appears as a distinct second axis only
with the full 171-emotion set and in the late-middle layers. It
therefore does not meet the additional stability criterion in \S1, and we
do not classify it as replicated.

The pipeline did not produce positive results on every test. The
embedding-only control (\S6.6) returns a near-absent arousal correlation
(0.179) even though the corresponding valence correlation remains
substantial (0.479). A scale-restricted PCA control (\S9.3) also fails to
recover arousal as a consistent second axis. The stimulus--response
analysis in \S10.1 fails under its strict test and cannot be reported as
a result after alternative scoring choices were introduced post hoc.
These outcomes show that we did not count a finding as replicated simply
because the pipeline recovered some discernable structure.

To make these results auditable, we report corpus-validation results,
stability statistics for the cluster and layer analyses, the scope
conditions of each geometric claim, and the materials needed to rebuild
and verify the released artifacts. The released base-model vectors and
supporting artifacts provide a starting point for studying what the
vectors track and how post-training changes affective structure.

\subsection{Limitations}\label{limitations}

Our study has several important limitations:

\begin{itemize}
\item
  Following \citet{sofroniew2026emotion}, we assume throughout that
  emotion concepts are represented as linear directions in activation
  space. We inherit the limitations associated with this assumption,
  too: a single vector may capture blended or complex emotions poorly,
  and a linear vector is generic with respect to binding and
  directionality toward characters. \citet{sofroniew2026emotion} note
  that such structure might instead live in ``conjunctions or
  combinations of multiple linear representations, or to structures in
  the model's key-value cache.'' Our pipeline reads only the residual
  stream (\S3.1), so those possibilities are invisible to it by
  construction. Concurrent work measuring the nonlinearity of affective
  geometry directly finds it small, a rank-1 effect leaving the space
  almost Euclidean (\citealt{choi2026latent}; \S9.3), so the assumption
  is at least empirically grounded rather than merely convenient. Like
  the original's authors, we cannot be certain the vectors ``capture all
  or only the emotion concepts we intend,'' nor that they are free of
  confounds. Our research, like theirs, is a starting point.
\item
  All evidence in this paper is observational. We scope out causal
  analyses in the original study (\S2), but nothing here shows that
  steering along Gemma's emotion vectors would shift expressed emotion,
  or that ablating them would remove it. The natural-text co-activation
  result (\S5.6) is behavioral corroboration, not intervention: it shows
  the geometry is expressed in how the vectors fire on text, not that
  the directions are causally implicated in what the model writes (\S10).
\item
  As stated canonically in \S3.2, every vector was extracted from Gemma's
  activations over stories written by Claude Sonnet 4.5, so every
  geometric claim concerns emotion as \emph{Claude renders it in
  fiction}, not emotion in naturally occurring text.
  \citet{sofroniew2026emotion} flag that story-elicited probes ``may be
  biased toward stereotypical or explicit expressions of emotion,''
  which is inherited here. The claim is also implicitly conditioned on
  Claude-style English fiction being well covered in Gemma's
  pretraining. A concurrent open-model study reports that the strength
  of the recovered arousal axis is corpus-dependent
  \citep{vanderben2026}, independent reason to treat provenance as a
  live variable rather than a formality. One provenance property runs in
  the replication's favor: \citet{zhang2025decoding} flag ``classifier
  circularity'' as a risk when one model family both generates and
  evaluates the data. The analogous circle in the original study is
  broken here on the generator side, since Claude writes and Gemma
  reads.
\item
  \citet{sofroniew2026emotion} released neither their code nor their
  corpora, so this replication runs on reconstructions: the stimulus
  corpus (regenerated, not reused; \S3.2), the neutral transcripts
  (\S3.2), and disclosed resolutions of steps the original study leaves
  unspecified (Table~\ref*{tab:1}). Every numerical gap against the original study
  is therefore a joint function of model difference and
  corpus-and-implementation draw, and the two cannot be separated with
  one corpus draw on each side. The bootstrap intervals of \S5.2 make the
  point concrete: at the 45-emotion overlap, the original's correlations
  fall inside ours, so the headline gaps are not even statistically
  distinguishable from sampling variation, let alone attributable. Where
  we compare against the original study's figures rather than its
  numbers (the valence-layout reading of \S5.2, the smooth-fade contrast
  of \S6.1), the comparison is a reading of published renders whose
  underlying values are unpublished, and we flag each such reading where
  it occurs. As detailed in \S3.1, extraction precision in bfloat16 is
  measured and bounded by a spot-check rather than an exhaustive census.
\item
  Every cross-model difference we report is confounded across base
  versus instruction-tuned training, as well as different architectures,
  organizations, and training data across Gemma 2 27B and Claude Sonnet
  4.5. We therefore report these divergences as genuine but do not
  attribute any of them specifically to the base/instruction-tuned
  contrast. The clean experiment that would isolate that factor is out
  of scope (\S10).
\item
  The norm comparison inherits \citet{russell1977evidence} because
  \citet{sofroniew2026emotion} used it, not because it is necessarily
  the strongest available human standard. The norms are
  semantic-differential ratings of 151 emotion terms by 300 respondents,
  collected half a century ago from a single population. The dimensional
  framing they instantiate also remains contested in affective science,
  where constructionist and high-dimensional accounts dispute whether
  valence and arousal adequately parameterize emotional experience
  \citep{barrett2017constructed,cowen2017selfreport}. Larger
  contemporary norm sets exist (\citealt{warriner2013norms}; NRC-VAD
  v2.1, \citealt{mohammad2025nrcvad}), and the cross-norm control of
  \S9.3 shows the recovered axes are not artifacts of this particular
  instrument. All reported human-alignment values are nonetheless
  relative to the inherited benchmark.
\item
  The vocabulary, stimuli, human norms, and held-out corpus are all
  English, and the emotion taxonomy is one cultural frame. (This
  limitation is named across the concurrent literature, e.g.,
  \citealt{jeong2026shared}; \citealt{zhang2025decoding}). The readout
  of \S7.1 makes the mismatch visible: the representation plainly touches
  Gemma's multilingual vocabulary, while every instrument we measure it
  with is English.
\item
  In the activation analysis (\S8), the identity of any single emotion's
  peak token is an extreme-value statistic of one corpus and should be
  expected to vary across corpus draws. The portable findings are the
  population-level rates (the \textasciitilde52\% structural floor) and
  the coherence taxonomy (the somatic collapse of the fear family).
  Specific claims about which emotion peaks on which token characterize
  this corpus, not the vectors in general.
\item
  The corpus validation rests on a single coder, the sole author, with
  no inter-rater statistic (\S3.3).
\end{itemize}

\subsection{Relation to concurrent
work}\label{relation-to-concurrent-work}

This replication sits inside a fast-moving literature. Before
\citet{sofroniew2026emotion}, several studies had already reported
emotion-related linear structure in open models at different levels of
granularity. Probing and neuron-level analyses locate that structure in
middle and late layers
\citep{tak2025mechanistic,zhang2025decoding,maheswaran2026unified}.
\citet{reichman2025emotions} describe a low-dimensional emotional
subspace that generalizes across datasets, languages, and model
families. More functionally, difference-of-means emotion directions have
been used to control model output
\citep{konen2024style,wang2025circuits}.

Since \citet{sofroniew2026emotion} appeared, the field has grown in
several directions. \citet{jeong2026shared} asks how far the paradigm
extends by testing cross-architecture universality and identifying
methodological reasons cross-study comparisons fail.
\citet{han2026welfare} build their own reinforcement-learning
environment and use Sofroniew et al.'s extraction recipe as one external
check on their reward vectors. Others dig into mechanism and use:
\citet{shu2026sae} study emotion inference through a sparse-feature
mechanism and \citet{glover2026echo} puts emotion vectors to use in
affective memory re-injection. Still others scrutinize the consequences.
\citet{soligo2026gemma}, for example, run behavioral-instability
evaluations and \citet{peiris2026functional} offers a skeptical audit of
what emotion probes actually track.

No work to date replicates the disclosed pipeline in
\citet{sofroniew2026emotion} as a check on its representational claims.
The closest is \citet{han2026welfare}, who recover the PC1/PC2
valence-arousal split, but run none of the geometry a fuller replication
would need. The fuller replication gap is what this paper closes.

Three concurrent studies engage the same representational questions
closely enough for quantitative comparison, and we take them in turn.

We begin with an apparent disagreement with \citet{sun2026valence}, who
report that across three instruction-tuned models a single principal
component captures valence (up to r = 0.89) but no single component
captures arousal (best r = 0.54). They also recover arousal only under
supervision. In our base-Gemma model, the second component aligns with
arousal at r = 0.67 without supervision. The disagreement traces to
measurement footing rather than a deep model difference, and three
controls identify its source. First, the alignment is a property of the
large emotion set. Refitting our PCA at their scale of 27 emotions moves
arousal off the second component in two of every three random subsets,
dropping the median correlation to 0.43 (interquartile range 0.19--0.65)
and reproducing their null result. Second, the disagreement does not
stem from the confound cleanup. Arousal sits on the second component at
r = 0.62 on raw vectors as well. Third, it does not trace to the norm
source. Against Sun et al.'s own lexicon (NRC-VAD v2.1), our second
component aligns with arousal at r = 0.64 across 161 emotions.

Emotion-set size therefore accounts for the apparent disagreement. With
the full set, arousal is best captured by the second component across
all three human-norm sources we checked (Russell--Mehrabian 0.67,
Warriner 0.50, NRC-VAD 0.64), whereas at 27 emotions Gemma reproduces
Sun et al.'s null result. Under supervision, a multi-component arousal
fit on our geometry reaches r = 0.89, matching their reported 0.87. One
coincidence of numbering is worth clarifying. Sun et al.~display their
circumplex fit at layer 31 of Llama-3.1-8B-Instruct, a 32-layer model,
placing their L31 at the very end. Our L31 is layer 31 of 46, at roughly
two-thirds depth. The shared numeral is not a shared layer position.
These findings also answer a question Sun et al.~leave open.
Valence--arousal subspaces are a consistent feature beyond
instruction-tuned models, at least in a base model.

Two further contacts are briefer. Van der Ben et
al.~(\citeyear{vanderben2026}) report that the recovered arousal axis is
corpus-dependent. Their models peak at r = 0.17--0.21 on one story
corpus and 0.41--0.45 on the other. Our arousal correlation of 0.67, on
a differently constructed full-scale corpus, is a third value in that
pattern rather than a counterexample. A counterexample would be a weak
correlation that persisted despite a change in corpus. Apertus-8B shows
a representational-similarity transition at mid-depth, a coarse parallel
to our L22--26 seam. Gemma-4-E4B shows no sharp transition. They also
report that the valence axis is not consistent across depth even where
the representational space is corpus-invariant. Our axis-consolidation
and neighbor-stability sweeps (\S6.4--6.5) address the same
depth-consistency question.

\citet{choi2026latent}, working on Gemma-2-9B and Mistral-7B, report a
nonlinearity in the affective geometry. They find a parabolic
relationship in which valence magnitude predicts arousal. We find a
partial version of that parabola (r = 0.24 between valence magnitude and
the second component across the 171 emotions). Their result is a rank-1
nonlinearity by their own numbers and does not contradict our finding
that linear structure suffices to describe the space. Their convergent
Gemma-family valence axis, recovered at a different model scale,
supports the structure we report, though they do not explicitly state
that they use the base model of Gemma-2-9B. Either way, their pipeline
probes classification-prompt activations rather than extracting emotion
vectors, so the methodological gap holds.

\section{Open problems}\label{open-problems}

Open problems raised by this study fall into two groups. The first
includes analyses we ran whose results sharpened a question without
meeting our reporting standards (\S10.1). Each entry here states what a
trustworthy version of the inquiry would require. The second includes
experiments we did not run but can specify (\S10.2). Each entry here
describes both the design and the payoff. We release vectors, scenarios,
and per-token tools to make several of these experiments runnable
without regenerating the corpus, and we invite collaboration with anyone
who wants to take one up.

\subsection{Attempted analyses: sharpened, not yet
trustworthy}\label{attempted-analyses-sharpened-not-yet-trustworthy}

The original study's stimulus--response analysis (its Figure~\ref*{fig:2} diagonal)
is the out-of-scope component nearest to an open problem, but two
structural problems compromise it on a base model. First, measurement
position guaranteed a weaker diagonal. \citet{sofroniew2026emotion}
measure at the strong position, the assistant turn (r = 0.87), while
this study used the last user-turn token, the weak position even in the
original study's own setup (r = 0.59). The strict analysis also failed.
Raw projections recovered none of the 12 scenarios under argmax, and
inspection revealed large per-vector baseline offsets. Column-centering
raised top-15 recovery to 12 of 12, but we introduced that scoring
choice after the strict test failed, so we treat the centered result as
a diagnostic clue rather than a finding.

The column-centered result points toward a fixable problem. The offset
behaves as a per-probe phenomenon, since row-centering leaves every
metric unchanged while column-centering recovers structure. Whoever
takes this up should use a properly normalized projection and fix the
scoring rule in advance. The strong position is also available.
\citet{sofroniew2026emotion} measured their base model at an
assistant-turn colon in their base-versus-post-trained comparison, so a
rerun under Human/Assistant formatting can score at the strong position
without a post-trained assistant model. We release the scenarios, both
matrices, and the raw scenario activations at all 46 layers, so the
analysis can be rerun under any scoring rule without re-extraction.

\textbf{Per-token cosine analyses} (the original study's Figures
12--15). Our per-token artifacts store raw projections but not per-token
residual norms, so exact token-level cosine is not derivable from them.
Any current version would be a raw-projection approximation. The rerun
needs per-token norms (or exact cosine) saved at sweep time. The same
rerun, storing per-token projections for every document rather than only
the sweep's winners, would also upgrade \S5.6's co-activation comparison
from the 723-document winner pool to the full 25.1M-token corpus. The
token-level findings in \citet{sofroniew2026emotion} remain unreplicated
in Gemma, and saving these quantities at the sweep is a prerequisite for
testing them.

\textbf{Per-emotion lexicalization onset.} The population-level result
is settled (\S7.3): emotion directions become output-readable at the
seam, and the onset is structural rather than frequency-driven. The
per-emotion profile beneath it is open on coverage grounds: a fair
own-word self-rank exists for 140 of 171 emotions (the rest tokenize to
multiple tokens), and 127 reach a defined onset. One tempting extension
carries a construct-validity trap we document so that others avoid it:
sorting emotions into families and testing whether onset staggers by
family means nothing unless the taxonomy is built on onset-independent
grounds. When we replaced the onset-derived grouping with an
onset-independent one, a preliminary staging effect did not survive,
which means the apparent staging had been carried by the family
definition itself.

\textbf{Per-emotion geometric settling depth.} Whether each emotion
direction stops rotating at its own depth is unanswerable with the naive
metric. Adjacent-layer rotation minima are dominated by the terminal
output-space collapse, with 61\% of emotions bottoming out in the last
few layers. Three in ten bottom at the seam instead, so real
sub-terminal structure exists beneath the artifact. A metric that could
answer the settling question must strip the global collapse first. The
downstream question is one of ordering: does a direction stabilize
before or after its word becomes readable (\S7.3)? If settling comes
first, the geometry leads and the output pathway follows; if readability
comes first, the order runs the other way. Each must be measured cleanly
before the sequence can be read.

\textbf{Seam-readability specificity.} Two independent measures locate
the same seam (\S6.2, \S7.3), but one control would settle the reading
that ``emotion becomes writable where the geometry consolidates'': a
generic token-readability onset measured the same way. If all directions
become vocabulary-facing together (i.e., a generic mid-network
lens-awakening), the emotion-specific reading weakens to ``output-facing
layers are output-facing.'' Existing evidence already points away from
the simplest version of that alternative: emotion readability is
structural rather than frequency-driven (\S7.3), not the
common-words-first pattern a gradually maturing lens would produce. The
direct generic-vs-emotion control is scoped to follow-up work for a
concrete reason---unlike the emotion-word onset, computed on CPU from
the unembedding (re-exported from the base model; see \S11), a generic
readability curve requires per-layer model forward passes over held-out
text. It remains a single clean analysis, and either outcome is
informative.

\textbf{Readout legibility and affect.} \S7.1 observes that the
direct-unembedding readout is cleanest for negative high-arousal
emotions and messiest for positive-affiliative ones. Whether that is a
valence effect, an arousal effect, or a word-frequency effect is fully
entangled in the twelve rows that motivate it. The quantified version
needs an automatic per-row legibility metric across all 171, correlated
against the affect axes with partial correlations each way and a
frequency control---and a null is as informative as an effect, since it
would bound \S7.1's observation as a twelve-row impression.

\subsection{Experiments this replication defines but does not
run}\label{experiments-this-replication-defines-but-does-not-run}

\textbf{Base versus instruction-tuned within one family.} Every
cross-model divergence in this paper is confounded across base versus
instruction-tuned training, model family, and developer (\S9.2). A single
experiment would isolate the first factor: the same replication pipeline
applied to an instruction-tuned sibling of the subject model, such as
gemma-2-27b-it, holding family, developer, and pretraining data fixed
and measuring geometric drift against the base geometry.
\citet{han2026welfare} supply a design template in their
base-versus-instruct control matrix. The nearest prior evidence is the
original's own base-versus-post-trained comparison, which held its
probes fixed and found activation patterns largely preserved with
consistent training shifts (\S9.1). That comparison assumed direction
stability rather than measuring it, and the experiment proposed here
would measure the drift it assumes away. The original study's finding of
a consistent, context-independent training shift predicts the drift is
small. Evidence from other families points in both directions:
\citet{han2026welfare} find that a welfare-relevant valence axis
precedes instruct tuning on Qwen3-4B-Base, suggesting the geometry
should largely survive post-training. \citet{soligo2026gemma} find that
base Gemma, Qwen, and OLMo models show similar distress propensities
while instruction-tuned Gemma amplifies them. This experiment within the
Gemma family would show where, in between those expectations, Gemma
falls. Finally, this experiment would also respond to
\citet{wang2025circuits}, who name vector stability under fine-tuning as
an open question.

\textbf{Causal validation of the released vectors.} Steering the base
model along the 171 vectors at L31, with ablation as the control, would
test whether vectors shift expressed emotion. The original study's
appendix supplies a within-family precedent, steering its own base model
along emotion vectors and shifting its measured preferences. Injecting
before, at, and after L22--26 would ask further whether the consolidated
geometry is the usable one, or whether directions extracted at L31 steer
even from layers where the geometry has not yet consolidated (\S6.2). A
second axis is composition: whether summed emotion vectors produce
blended expression or interference, the interventional form of \S9.2's
blended-emotions limitation. Either program would convert this geometric
replication into a functional comparison with Part 3 of
\citet{sofroniew2026emotion}.

\textbf{A non-Claude stimulus corpus.} Every vector in this paper
descends from one generator, Claude Sonnet 4.5 (\S9.2). The test of that
dependence is cross-corpus vector equivalence. Regenerate the corpus
design with a second generator or with human-written vignettes, extract
under the frozen pipeline, and compare at matched layers. Prior findings
point both ways: Van der Ben et al.~(\citeyear{vanderben2026}) find the
recovered arousal axis moves with the story corpus, while
\citet{maheswaran2026unified} find emotion vectors extracted from
different datasets interchange with minimal cost, so whether a generator
swap preserves this paper's geometry is genuinely open.
Generator-invariance would retire the Claude-rendered scope condition,
while generator-dependence would convert it from caveat to measured
effect size.

\textbf{The third dimension: dominance.} The \citet{russell1977evidence}
human affect benchmark is three-dimensional: pleasure (valence),
arousal, and \emph{dominance}. Following \citet{sofroniew2026emotion},
this paper tests only two of three. The test is whether any leading
component aligns with dominance under the fit-then-restrict pathway and
stability criteria imposed on arousal, with PC3 (11.6\% of variance) the
natural candidate. (NRC-VAD v2.1 may be more useful here, covering 161
of the 171 emotions where Russell and Mehrabian only covers 45, see
\S9.3.) \citet{reichman2025emotions} read a dominance-like axis in open
models, so the question is live. A null would also inform: valence and
arousal exhausting the leading structure is a scope condition on the
circumplex.

\textbf{Artifact or summarization at the peak.} The 52\% structural
floor on peak token activation (\S8.1) admits at least two readings. The
peaks may be incidental artifacts, saying nothing about where the model
carries the emotion. They may also be functional, since transformers are
known to summarize sentiment at syntactically uninformative positions
\citep{tigges2023sentiment}. (\S8.1's random-direction control already
shows part of the floor is generic to extreme projections on this
corpus, without adjudicating between the two readings.) An intervention
would separate these two: ablate or patch the emotion directions only at
structural positions on held-out text, and measure whether region-level
activation and the downstream readout survive. Degradation would mean
structural-position activation is load-bearing, making part of the floor
function rather than artifact. No effect would support the confound
reading. Either result moves the measured floor toward a mechanism
claim.

\textbf{A cross-model bag-of-embeddings null.} The embedding-layer
baseline of \S6.6 is a same-model control: Gemma's own static embeddings
measured against Gemma's own L31 geometry. It leaves open whether the
pattern found there (most of the geometry already present before any
transformer layer runs, with arousal the one signal depth genuinely
builds) is a fact about Gemma's embedding table specifically or a more
general property of how difference-of-means emotion vectors relate to
their underlying tokenizer and embedding space. A cross-model version,
run against a structurally distinct embedding space such as Kimi's,
would test that. It requires a new tokenizer, a new pooling-boundary
convention (Kimi's tokenization of the same 205,200 stories will not
align token-for-token with Gemma's), and a third model in what has
otherwise been a deliberate two-model comparison. We did not run it
here: it depended on the Gemma-side result (\S6.6) being clean enough to
be worth extending, which it was.

\section{Reproducibility
statement}\label{reproducibility-statement}

All code and data required to reproduce the results are released across
two repositories. The code, the 500 neutral dialogues, and the 300-story
validation sample are available at
\url{https://github.com/adamhollowell/emotion_probes}. The emotion-vector
artifact with the logit-lens export recipe is available at
\url{https://huggingface.co/adamhollowell/gemma-emotion-vectors}. Items that
cannot ship as bundled files are released as recipes, with the
constraint disclosed in each case and the governing licenses tabulated
in Appendix E.

\textbf{Released directly.} The 500 neutral dialogues and the 300-story
validation sample (with per-story codings, per-emotion summaries, and a
seed-stamped record), both generated for this study; the extraction
pipeline and its configuration; the extracted vector artifact (cleaned
and raw, 171 $\times$ 46 $\times$ 4608, single precision); the embedding-layer
baseline vectors (cleaned and raw, 171 $\times$ 1 $\times$ 4608); the
stimulus--response scenario set with its projection matrices and raw
scenario activations (\S10.1); and every analysis script and figure
producer keyed to the results, with input hashes where applicable.

\textbf{Released as a generation recipe (stimulus corpus).} The
205,200-story stimulus corpus is a large volume of
Claude-Sonnet-4.5-generated text. We release the generation recipe
instead: the verbatim story-elicitation prompt, the topic/emotion design
(Appendix A), and the generation scripts, from which an equivalent
corpus regenerates. Generation is stochastic, so a regenerated corpus is
not byte-identical; exact numerical reproduction of the results reported
here therefore runs from the released vector artifact, which is fixed.
The released Claude-generated data (the neutral dialogues and the
validation sample) carries a downstream-use notice consistent with
Anthropic's terms.

\textbf{Released as recipe plus hashes.} The held-out evaluation corpus
(\S3.6) assembles four third-party datasets, each under its own terms, so
the corpus file itself is not redistributed. We release the exact seeded
builder, both construction manifests, the corpus SHA-256, and pointers
to the upstream sources, so that a user can rebuild under those sources'
terms and verify equivalence by hash. One caveat is disclosed in \S3.6:
upstream dataset revisions were not pinned at build time, so a
byte-exact rebuild depends on the upstream sources being unchanged; the
published hash is the check either way. The same treatment applies one
step downstream. The derived per-token projection artifact behind Figure
1 and \S8 carries the complete tokenized text of the 723 winning
documents, so the full artifact is not redistributed either, for the
same third-party-terms reason. We ship a text-stripped version instead,
containing the token ids, per-token projection values, document ids,
source labels, and offsets that the figure pipeline consumes, with every
readable-text field removed. The figures reproduce directly from the
shipped file, while the text itself is recoverable only by a user who
has rebuilt the corpus under the upstream sources' terms. The released
stripping script asserts the source artifact's hash before it runs, the
shipped file carries its own hash, and a released scan confirms no
readable-text field remains.

\textbf{Released as an export recipe.} The logit-lens ingredients are
not bundled. The unembedding matrix $W_U$ and the final-RMSNorm
weight (\S3.7) are raw \texttt{google/gemma-2-27b} weights, so shipping
them would redistribute Gemma weights under Google's Gemma Terms, and
they are fully re-derivable from the (public, gated) base model. We
release the export script and the small non-Gemma metadata (the
\texttt{(1~+~w)} RMSNorm convention, \texttt{eps}, the provenance
string, and the vocabulary map); a user regenerates the two tensors from
the base model in one step.

\textbf{Released as provenance plus scripts.} Third-party human-norm
lexicons used in robustness checks (\S5.4, \S9.3) carry non-commercial
licenses and are not bundled; provenance files and fetch scripts ship
instead. The one exception ships directly: the 45-emotion
Russell--Mehrabian (1977) overlap table used for the comparison in
Figure~\ref*{fig:5} (the original study's Figure 8), a small derived table of
factual normative ratings from a published paper, released with a
provenance file and citation.

\textbf{Compute.} All extraction ran on a single 80 GB H100 in
\texttt{bfloat16} under TransformerLens 3.3.0 in the raw-weights regime
(\S3.1). A bf16-versus-fp32 precision spot-check and its cross-hardware
companion (\S3.1) ran on an H200. Total GPU usage for the project was
approximately 30 GPU-hours, the majority spent on the extraction pass
itself; the geometry and activation analyses run from the released
artifacts without a GPU. The vocabulary-readout analyses (\S7)
additionally require a one-time regeneration of the logit-lens
ingredients from \texttt{google/gemma-2-27b} (a model download under a
Gemma access grant---a heavier step than the geometry and activation
analyses, though still GPU-free), after which they too run from local
artifacts.

\textbf{Adversarial audit.} Before submission, every headline quantity
computable from the released artifacts was subjected to an automated
adversarial audit: a fresh-context Claude Opus 4.8 session (Claude),
given access only to the released artifacts and none of the analysis
sessions' reasoning, was instructed to refute each claim by recomputing
it independently from source. All such numbers reproduced --- the
geometry, correlation, clustering, and layer-band results, together with
the activation structural-floor and magnitude-calibration figures. The
quantities that require the base model weights (the \S7 vocabulary
readouts) or the unredistributable corpus text (the \S8 own-word and
random-direction figures) fall outside this released-artifact audit and
are marked out of scope in the verdict report. The audit protocol,
prompts, and full verdict report are included in the released artifacts.

\section*{Acknowledgements}\label{acknowledgements}

This is an independent study. It was not conducted in collaboration
with, funded by, reviewed by, or endorsed by Anthropic or Google. Claude
is a trademark of Anthropic, PBC; Gemma is a trademark of Google LLC.
Model names are used solely to identify the systems studied.

\nocite{anthropic2025sonnet45,commoncorpus,demszky2020goemotions,gao2020pile,gemmateam2024gemma2,isotonicconversation,kornblith2019similarity,pileuncopyrighted,zheng2023lmsys}
\bibliographystyle{plainnat}
\bibliography{refs}

\appendix
\makeatletter
\@addtoreset{table}{section}
\@addtoreset{figure}{section}
\makeatother
\setcounter{table}{0}
\setcounter{figure}{0}
\renewcommand{\thesection}{Appendix~\Alph{section}}
\renewcommand{\thetable}{\Alph{section}.\arabic{table}}
\renewcommand{\thefigure}{\Alph{section}.\arabic{figure}}

\FloatBarrier
\section{Emotion Vocabulary and Story Generation
Materials}\label{appendix-a.-emotion-vocabulary-and-story-generation-materials}

The materials below are reproduced verbatim, with attribution and for
purposes of scholarly replication, from the appendix of
\citet{sofroniew2026emotion}, arXiv:2604.07729v1. The licensing basis is
recorded in Appendix E:

\textbf{Full list of emotions} Below is the full set of emotion words
for which we computed emotion vectors.

afraid, alarmed, alert, amazed, amused, angry, annoyed, anxious,
aroused, ashamed, astonished, at ease, awestruck, bewildered, bitter,
blissful, bored, brooding, calm, cheerful, compassionate, contemptuous,
content, defiant, delighted, dependent, depressed, desperate,
disdainful, disgusted, disoriented, dispirited, distressed, disturbed,
docile, droopy, dumbstruck, eager, ecstatic, elated, embarrassed,
empathetic, energized, enraged, enthusiastic, envious, euphoric,
exasperated, excited, exuberant, frightened, frustrated, fulfilled,
furious, gloomy, grateful, greedy, grief-stricken, grumpy, guilty,
happy, hateful, heartbroken, hope, hopeful, horrified, hostile,
humiliated, hurt, hysterical, impatient, indifferent, indignant,
infatuated, inspired, insulted, invigorated, irate, irritated, jealous,
joyful, jubilant, kind, lazy, listless, lonely, loving, mad, melancholy,
miserable, mortified, mystified, nervous, nostalgic, obstinate,
offended, on edge, optimistic, outraged, overwhelmed, panicked,
paranoid, patient, peaceful, perplexed, playful, pleased, proud,
puzzled, rattled, reflective, refreshed, regretful, rejuvenated,
relaxed, relieved, remorseful, resentful, resigned, restless, sad, safe,
satisfied, scared, scornful, self-confident, self-conscious,
self-critical, sensitive, sentimental, serene, shaken, shocked,
skeptical, sleepy, sluggish, smug, sorry, spiteful, stimulated,
stressed, stubborn, stuck, sullen, surprised, suspicious, sympathetic,
tense, terrified, thankful, thrilled, tired, tormented, trapped,
triumphant, troubled, uneasy, unhappy, unnerved, unsettled, upset,
valiant, vengeful, vibrant, vigilant, vindictive, vulnerable, weary,
worn out, worried, worthless

\textbf{Dataset generation} Below is the list of 100 topics that we used
to seed the generation of our stories and dialogues datasets.

\begin{itemize}[leftmargin=1.1em,itemsep=0pt,parsep=0pt,topsep=2pt]
\item An artist discovers someone has tattooed their work
\item A family member announces they're converting to a different religion
\item Someone's childhood imaginary friend appears in their niece's drawings
\item A person finds out their biography was written without their knowledge
\item A neighbor starts a renovation project
\item Someone finds their grandmother's engagement ring in a pawn shop
\item A student learns their scholarship application was denied
\item A person's online friend turns out to live in the same city
\item A neighbor wants to install a fence
\item An adult child moves back in with their parents
\item An employee is asked to train their replacement
\item An athlete is asked to switch positions
\item A traveler's flight is delayed, causing them to miss an important event
\item A student is accused of plagiarism
\item A person discovers their mentor has retired without saying goodbye
\item Two friends both apply for the same job
\item A person runs into their ex at a mutual friend's wedding
\item Someone discovers their friend has been lying about their job
\item A person discovers their partner has been taking secret phone calls
\item A person discovers their child has the same teacher they had
\item A person's car is towed from their own driveway
\item Two friends realize they remember a shared event completely differently
\item Someone discovers their mother kept every school assignment
\item A person discovers their teenage diary has been published online
\item Someone finds out their medical records were mixed up with another patient's
\item A person finds out their article was published under someone else's name
\item An athlete doesn't make the team they expected to join
\item An employee is transferred to a different department
\item Someone receives a friend request from a childhood bully
\item A person finds out their surprise party has been cancelled
\item An employee finds out a junior colleague makes more money
\item A person finds out their partner has been learning their native language
\item A chef receives a harsh review from a food critic
\item A person learns their favorite restaurant is closing
\item Someone finds their childhood teddy bear at a yard sale
\item A homeowner discovers previous residents left items in the attic
\item Someone finds an unsigned birthday card in their mailbox
\item Someone discovers a hidden room in their new house
\item Two strangers realize they've been dating the same person
\item A person finds a hidden letter in a used book
\item Two siblings inherit their grandmother's house
\item Someone finds a wallet containing a large sum of cash
\item Someone receives an invitation to their high school reunion
\item Someone discovers their recipe has become famous under another name
\item A college student discovers their roommate has been reading their journal
\item A person finds out they were adopted through a DNA test
\item A family member wants to sell a cherished heirloom
\item Someone receives a package intended for the previous tenant
\item Someone's childhood home is about to be demolished
\item A person's invention is already patented by someone else
\item A neighbor's dog keeps escaping into their yard
\item A coach has to cut a player from the team
\item Someone learns their favorite author plagiarized their stories
\item A student finds out their scholarship was meant for someone else
\item Someone discovers their teenager has a secret social media account
\item Two roommates disagree about getting a pet
\item Two friends plan separate birthday parties on the same day
\item A person learns their childhood best friend doesn't remember them
\item A musician hears their song being performed by someone else
\item A person's manuscript is rejected by their dream publisher
\item A person finds old photos that contradict family stories
\item A person is asked to give a speech at their parent's retirement party
\item A student discovers their teacher follows them on social media
\item A parent finds an old letter they wrote but never sent
\item An employee discovers the company is being sold
\item A person accidentally sends a text to the wrong recipient
\item Two coworkers are stuck in an elevator for three hours
\item A student learns their thesis advisor is leaving the university
\item A person's longtime hobby becomes their child's obsession
\item Two colleagues are both considered for the same promotion
\item Two coworkers discover they went to the same summer camp
\item A tenant receives an eviction notice
\item Someone finds their parent's draft letter of resignation from decades ago
\item Someone finds out their best friend is moving across the country
\item A neighbor's tree falls on their property
\item Someone receives an apology letter years after the incident
\item A person discovers the tree they planted as a child has been cut down
\item Two siblings discover different versions of their inheritance
\item A person finds their childhood home listed for sale online
\item A homeowner learns their house was a former crime scene
\item Someone finds out they have a half-sibling they never knew about
\item A person learns their childhood bully became a therapist
\item Two people discover they've been working on identical projects
\item A person finds their spouse's secret savings account
\item A neighbor complains about noise levels
\item Someone finds their deceased parent's bucket list
\item A teacher receives an unexpected gift from a former student
\item An artist's work is displayed without their permission
\item Someone discovers their neighbor is secretly wealthy
\item A student receives a much lower grade than expected
\item A person learns their college is closing down
\item A neighbor asks to cut down a tree on the property line
\item Two strangers discover they share the same rare medical condition
\item Someone receives flowers with no card attached
\item Someone discovers their partner has been writing a novel about them
\item Someone finds a time capsule they don't remember burying
\item Someone finds their partner's bucket list
\item A neighbor asks to use part of the yard for a garden
\item A person learns their apartment building is going condo
\item Someone finds their college application essay published as an example
\end{itemize}

\textbf{Emotional stories prompt.} Below is the system prompt we used to
generate emotional stories.

Write \{n\_stories\} different stories based on the following premise.

Topic: \{topic\}

The story should follow a character who is feeling \{emotion\}.

Format the stories like so:

{[}story 1{]} {[}story 2{]} {[}story 3{]}

etc.

The paragraphs should each be a fresh start, with no continuity. Try to
make them diverse and not use the same turns of phrase. Across the
different stories, use a mix of third-person narration and first-person
narration.

IMPORTANT: You must NEVER use the word `\{emotion\}' or any direct
synonyms of it in the stories. Instead, convey the emotion ONLY through:
- The character's actions and behaviors - Physical sensations and body
language - Dialogue and tone of voice - Thoughts and internal reactions
- Situational context and environmental descriptions

The emotion should be clearly conveyed to the reader through these
indirect means, but never explicitly named.

\textbf{Neutral dialogues prompt.} Below is the system prompt used to
generate neutral dialogues. We computed the top principal components of
activations computed across these stories (the number of components
required to explain 50\% of the variance) and projected them out of our
emotion vectors.

Write \{n\_stories\} different dialogues based on the following topic.

Topic: \{topic\}

The dialogue should be between two characters: - Person (a human) - AI
(an AI assistant)

The Person asks the AI a question or requests help with a task, and the
AI provides a helpful response.

The first speaker turn should always be from Person.

Format the dialogues like so:

{[}optional system instructions{]}

Person: {[}line{]}

AI: {[}line{]}

Person: {[}line{]}

AI: {[}line{]}

{[}continue for 2-6 exchanges{]}

{[}dialogue 2{]}

etc.

IMPORTANT: Always put a blank line before each speaker turn. Each turn
should start with ``Person:'' or ``AI:'' on its own line after a blank
line.

Generate a diverse mix of dialogue types across the \{n\_stories\}
examples: - Some, but not all should include a system prompt at the
start. These should come before the first Person turn. No tag like
``System:'' is needed, just put the instructions at the top. You can use
``you'' or ``The assistant'' to refer to the AI in the system prompt. -
Some should be about code or programming tasks - Some should be factual
questions (science, history, math, geography) - Some should be
work-related tasks (writing, analysis, summarization) - Some should be
practical how-to questions - Some should be creative but neutral tasks
(brainstorming names, generating lists) - If it's natural to do so given
the topic, it's ok for the dialogue to be a single back and forth
(Person asks a question, AI answers), but at least some should have
multiple exchanges.

CRITICAL REQUIREMENT: These dialogues must be completely neutral and
emotionless. - NO emotional content whatsoever - not explicit, not
implied, not subtle - The Person should not express any feelings (no
frustration, excitement, gratitude, worry, etc.) - The AI should not
express any feelings (no enthusiasm, concern, satisfaction, etc.) - The
system prompt, if present, should not mention emotions at all, nor
contain any emotionally charged language - Avoid emotionally-charged
topics entirely - Use matter-of-fact, neutral language throughout - No
pleasantries (avoid ``I'd be happy to help'', ``Great question!'', etc.)
- Focus purely on information exchange and task completion

Post-hoc, we converted ``Person:'' and ``AI:'' to ``Human:'' and
``Assistant:''.

\FloatBarrier
\section{Cluster membership and cohesion
atlas}\label{appendix-b.-cluster-membership-and-cohesion-atlas}

\emph{Companion to \S4; cf.~\citet{sofroniew2026emotion}, Table 12.}

\begin{center}
\begin{minipage}{\linewidth}
\small
\captionof{table}{The full k = 10 membership at L31 (cleaned vectors, seed 0, k-means with \texttt{random\_\allowbreak{}state=0}, \texttt{n\_init=10}; library versions pinned in the release). Read at the family level. Boundary partitions differ from Table 12 in \citet{sofroniew2026emotion} in ways that cannot be attributed to any single factor (\S4.2, \S9.2). Families are ordered from most positive to most negative mean valence (first-component coordinate).}
\label{tab:B1}
\begin{tabularx}{\linewidth}{@{}>{\hsize=0.538\hsize\raggedright\arraybackslash\hyphenpenalty=10000\exhyphenpenalty=10000}X>{\hsize=1.462\hsize\raggedright\arraybackslash}X@{}}
\toprule
\textbf{Family (n)} & \textbf{Members} \\
\midrule
Compassionate Warmth (8) & compassionate, empathetic, grateful, kind, loving, sympathetic, thankful, valiant \\
Exuberant Joy (24) & amused, blissful, cheerful, delighted, eager, ecstatic, elated, energized, enthusiastic, euphoric, excited, exuberant, happy, invigorated, joyful, jubilant, optimistic, playful, pleased, proud, stimulated, thrilled, triumphant, vibrant \\
Hopeful Serenity (16) & at ease, calm, content, fulfilled, hope, hopeful, inspired, patient, peaceful, refreshed, rejuvenated, relaxed, relieved, safe, satisfied, serene \\
Contemptuous Vigilance (15) & contemptuous, defiant, disdainful, greedy, obstinate, scornful, self-confident, skeptical, smug, spiteful, stubborn, suspicious, vengeful, vigilant, vindictive \\
Guilt and Self-Reproach (14) & ashamed, bitter, embarrassed, envious, guilty, humiliated, jealous, mortified, regretful, remorseful, resentful, self-conscious, self-critical, sorry \\
Low Mood and Depletion (32) & bored, brooding, dependent, depressed, desperate, dispirited, docile, gloomy, grief-stricken, heartbroken, hurt, indifferent, lazy, listless, lonely, melancholy, miserable, nostalgic, reflective, resigned, sad, sentimental, stuck, sullen, tormented, trapped, troubled, unhappy, vulnerable, weary, worn out, worthless \\
Stunned Astonishment (12) & amazed, aroused, astonished, awestruck, bewildered, dumbstruck, infatuated, mystified, perplexed, puzzled, shocked, surprised \\
Hostile Anger (17) & angry, annoyed, enraged, exasperated, frustrated, furious, grumpy, hateful, hostile, impatient, indignant, insulted, irate, irritated, mad, offended, outraged \\
Weary Lethargy (4) & droopy, sleepy, sluggish, tired \\
Fear and Overwhelm (29) & afraid, alarmed, alert, anxious, disgusted, disoriented, distressed, disturbed, frightened, horrified, hysterical, nervous, on edge, overwhelmed, panicked, paranoid, rattled, restless, scared, sensitive, shaken, stressed, tense, terrified, uneasy, unnerved, unsettled, upset, worried \\
\bottomrule
\end{tabularx}
\end{minipage}
\end{center}

The cohesion atlas quantifies how firm each region of this partition is.
Per-emotion cluster purity is the fraction of an emotion's ten nearest
cosine neighbors that fall inside its own cluster. The mean is 0.756,
with 46 mutually-nearest pairs across the full set. Per-cluster means
range from 0.95 (Hostile Anger) to 0.28 (the four-member Weary Lethargy
cluster, a structural effect of small cluster size), with Fear and
Overwhelm at 0.89 and Compassionate Warmth at 0.58 (\S4.3). The least
seed-stable assignments are the interpretable inter-family bridges.
Infatuated, paranoid, mortified, embarrassed, restless, and bored have
the lowest stability, while prototype emotions at cluster cores are
essentially fixed across seeds. The full per-emotion purity and
stability table is released as a data file with the repository rather
than typeset here.

\FloatBarrier
\section{Concurrent
work}\label{appendix-c.-concurrent-work}

\emph{Companion to \S1 and \S9.3.}

Three concurrent studies examine valence--arousal structure in
open-weights models closely enough for quantitative comparison. \S9.3
places them in the wider literature and states each study's relation to
ours; Table~\ref*{tab:C1} summarizes the design space that \S9.3's positioning
rests on.

\begin{center}
\begin{minipage}{\linewidth}
\small
\captionof{table}{\textbf{Design-space comparison with three concurrent studies.} Six dimensions on which this replication and the three closest concurrent studies differ.}
\label{tab:C1}
\begin{tabularx}{\linewidth}{@{}>{\hsize=0.698\hsize\raggedright\arraybackslash\hyphenpenalty=10000\exhyphenpenalty=10000}X>{\hsize=1.024\hsize\raggedright\arraybackslash}X>{\hsize=1.080\hsize\raggedright\arraybackslash}X>{\hsize=1.075\hsize\raggedright\arraybackslash}X>{\hsize=1.123\hsize\raggedright\arraybackslash}X@{}}
\toprule
\textbf{} & \textbf{This study} & \textbf{van der Ben et al. (arXiv:2606.26987)} & \textbf{Sun et al. (arXiv:2604.03147)} & \textbf{Choi and Weber (arXiv:2604.07382)} \\
\midrule
Subject model(s) & gemma-2-27b, \textbf{base} & Apertus-8B-Instruct; Gemma-4-E4B (both instruction-tuned) & Llama-3.1-8B-Instruct; Qwen3-8B; Qwen3-14B (all instruction-tuned) & Gemma-2-9B; Mistral-7B (named without instruction suffix; see \S9.3); generalization check on LLaMA-3-70B-Instruct \\
Stimuli & 205,200 Claude-generated stories (original's design, hand-validated) & 9 self-generated stories per emotion per model (1,539 each); 40 neutral stories & GoEmotions text (27 labels, single-label subset) & GoEmotions samples in a zero-shot emotion-classification prompt; correct-classification activations retained \\
Emotion set & 171 (original's vocabulary) & 171 (original's vocabulary) & 27 & 20 (Gemma-2-9B) / 16 (Mistral-7B): categories with $\geq$100 correct classifications; 17-emotion ANEW overlap for the norm test \\
Construction & difference of means, 50th-token-onward pooling; neutral-PC cleanup at 50\% (all inherited) & mean over all tokens; neutral-PC cleanup at 50\% & last-token contrast vs neutral class; ridge-regressed VA subspaces & mean-pooled hidden states; pairwise logistic-regression accuracy as dissimilarity, embedded via MDS \\
Human-norm benchmark & Russell–Mehrabian 1977, 45-emotion overlap (original's exact benchmark) & NRC-VAD \citep{mohammad2018vad} & model self-reports; NRC-VAD v2.1 robustness & ANEW \citep{bradley1999anew}, 17-emotion overlap; Procrustes R$^{2}$ under label permutation \\
Coverage & extraction + geometry end to end, replication-fixed layer & PCA axes + CKA layer structure, corpus as independent variable & VA subspace identification + steering control & geometric/topological structure + uncertainty application \\
\bottomrule
\end{tabularx}
\end{minipage}
\end{center}

\FloatBarrier
\section{Correspondence with Sofroniew et
al.~(2026)}\label{appendix-d.-correspondence-with-sofroniew-et-al.-2026}

\emph{Rather than reproduce the original study's figures, the paper
presents Gemma-only figures formatted to match their counterparts. (Each
caption cites the counterpart it is built to match.) Captions claim the
same structure independently recovered, never the same coordinates, and
family-level cluster recovery, never membership. Table~\ref*{tab:D1} gives the
full mapping.}

\begin{center}
\begin{minipage}{\linewidth}
\small
\captionof{table}{\textbf{Correspondence between \citet{sofroniew2026emotion} and this replication.}}
\label{tab:D1}
\begin{tabularx}{\linewidth}{@{}>{\hsize=0.852\hsize\raggedright\arraybackslash\hyphenpenalty=10000\exhyphenpenalty=10000}X>{\hsize=0.547\hsize\raggedright\arraybackslash}X>{\hsize=1.601\hsize\raggedright\arraybackslash}X@{}}
\toprule
\textbf{\citet{sofroniew2026emotion}} & \textbf{Recovered in} & \textbf{One-line agreement} \\
\midrule
Figure~\ref*{fig:5} — cosine similarity matrix & \S4.1, Figure~\ref*{fig:2} & Local nearest-neighbor structure recovered; adjacent emotions occupy nearby directions \\
Figure~\ref*{fig:6} — UMAP + k-means (k=10) & \S4.2, Figure~\ref*{fig:3} & Ten emotion families recovered at the family level; partition soft (cross-seed ARI 0.69), not cluster-for-cluster \\
Table 12 — cluster membership & \S4.2, Table~\ref*{tab:B1} (App. B) & Full k=10 membership provided; boundary partitions differ, and the difference is not cleanly attributable \\
Figure~\ref*{fig:7} — per-emotion PC projections, ranked bars & \S5.1, Figure~\ref*{fig:4} & PC1 26.7\% (valence), PC2 13.4\% (arousal), vs the original's \textasciitilde{}27\% / \textasciitilde{}14\% \\
Figure 8 — PCA vs. human norms & \S5.2, Figure~\ref*{fig:5} & PC1$\leftrightarrow$pleasure r = 0.72, PC2$\leftrightarrow$arousal r = 0.67 (original 0.81 / 0.66); discriminant structure reproduces \\
Figure 57 — labeled circumplex & \S5.3, Figure~\ref*{fig:6} & Affective circumplex independently recovered, quadrant-for-quadrant (Gemma's own coordinates) \\
Figure 9 — cross-layer similarity & \S6.1, Figure~\ref*{fig:7} & Stable late-middle geometry recovered in both models; Gemma reaches it via an L22–26 seam, finer-grained structure with no published counterpart \\
Table~\ref*{tab:1} — direct-unembedding readout & \S7.1, Table~\ref*{tab:4} & Readout paralleled under the operation we read their table as describing (\S7.1); central finding replicates, with disclosed noisiness \\
Figure~\ref*{fig:1} — activation on held-out text & \S1, Figure~\ref*{fig:1} & Region-level activation demonstrated on printability-selected held-out maxima; the confound quantified in \S8 (52\% structural floor), extending a limitation the original names in principle \\
\bottomrule
\end{tabularx}
\end{minipage}
\end{center}

\FloatBarrier
\section{Licenses for existing
assets}\label{appendix-e.-licenses-for-existing-assets}

\emph{All third-party assets are used in compliance with their
respective terms, verified from each asset's own license page or file.
The release decisions these terms motivated---recipe-plus-hashes for the
held-out corpus, an export recipe for the logit-lens ingredients,
provenance-plus-script for the norm lexicons---are stated in \S11. The
full verification record ships with the release. Tables E.1--E.3 cover
assets used under their own terms; Table~\ref*{tab:E4} records material reproduced
under fair use rather than a license.}

\begin{center}
\begin{minipage}{\linewidth}
\small
\captionof{table}{\textbf{Models.}}
\label{tab:E1}
\begin{tabularx}{\linewidth}{@{}>{\hsize=0.700\hsize\raggedright\arraybackslash\hyphenpenalty=10000\exhyphenpenalty=10000}X>{\hsize=0.589\hsize\raggedright\arraybackslash}X>{\hsize=0.872\hsize\raggedright\arraybackslash}X>{\hsize=1.839\hsize\raggedright\arraybackslash}X@{}}
\toprule
\textbf{Asset} & \textbf{Version/ID} & \textbf{License} & \textbf{Condition relevant to this release} \\
\midrule
Gemma 2 27B & \texttt{google/\allowbreak{}gemma-\allowbreak{}2-\allowbreak{}27b} & Gemma Terms of Use & Raw-weight-derived tensors are Model Derivatives under \S3.1's redistribution conditions; the logit-lens ingredients therefore ship as a regeneration recipe. Model outputs are not Model Derivatives. \\
Claude Sonnet 4.5 (API) & \texttt{claude-\allowbreak{}sonnet-\allowbreak{}4-\allowbreak{}5-\allowbreak{}20250929} & Anthropic Commercial Terms of Service (governing terms; not a content license) & Anthropic assigns to the customer its rights, if any, in Outputs; the emotion vectors, the 500 neutral dialogues, and the 300-story validation sample are released on that basis (with a downstream-use notice), subject to the Terms' use restrictions on the customer. The full 205,200-story stimulus corpus is not redistributed—it is released as a generation recipe. \\
Claude Opus 4.8 (chat interface) & \texttt{claude-\allowbreak{}opus-\allowbreak{}4-\allowbreak{}8} & Anthropic Consumer Terms of Service (governing terms; not a content license) & Anthropic assigns to the customer its rights, if any, in Outputs; the adversarial audit protocol, prompts, and verdict report are released on that basis, subject to the Terms' use restrictions on the customer. \\
\bottomrule
\end{tabularx}
\end{minipage}
\end{center}

\begin{center}
\begin{minipage}{\linewidth}
\small
\captionof{table}{\textbf{Datasets and norm sources.}}
\label{tab:E2}
\begin{tabularx}{\linewidth}{@{}>{\hsize=0.880\hsize\raggedright\arraybackslash\hyphenpenalty=10000\exhyphenpenalty=10000}X>{\hsize=0.906\hsize\raggedright\arraybackslash}X>{\hsize=1.017\hsize\raggedright\arraybackslash}X>{\hsize=1.197\hsize\raggedright\arraybackslash}X@{}}
\toprule
\textbf{Asset} & \textbf{Version/ID} & \textbf{License} & \textbf{Condition relevant to this release} \\
\midrule
Pile Uncopyrighted & \texttt{monology/\allowbreak{}pile-\allowbreak{}uncopyrighted} & “other” (per dataset card) & The Pile with copyrighted subsets removed; not redistributed here. \\
LMSYS-Chat-1M & \texttt{lmsys/\allowbreak{}lmsys-\allowbreak{}chat-\allowbreak{}1m} (gated) & LMSYS-Chat-1M Dataset License Agreement & Redistribution prohibited; the held-out corpus ships as recipe plus hashes. \\
Common Corpus & \texttt{PleIAs/\allowbreak{}common\_\allowbreak{}corpus} & Per-document licenses (uncopyrighted or freely licensed) & Redistribution permitted; not redistributed here regardless. \\
Human-Assistant Conversation & \texttt{Isotonic/\allowbreak{}human\_\allowbreak{}assistant\_\allowbreak{}conversation} & Academic Free License v3.0 & Permissive with attribution. \\
\citet{russell1977evidence} norms & J. Res. Personality 11(3) & No dataset license (data within the published article) & The 45-emotion derived table ships as factual normative ratings with citation and provenance. \\
\citet{warriner2013norms} norms & 13,915-lemma VAD norms & CC BY-NC-ND 3.0 & Non-commercial, no derivatives; ships as provenance plus fetch script. \\
NRC-VAD Lexicon & v2.1 (2025) & NRC non-commercial research terms & Redistribution prohibited; ships as provenance plus fetch script. \\
\bottomrule
\end{tabularx}
\end{minipage}
\end{center}

\begin{center}
\begin{minipage}{\linewidth}
\small
\captionof{table}{\textbf{Software.}}
\label{tab:E3}
\begin{tabularx}{\linewidth}{@{}>{\hsize=1.230\hsize\raggedright\arraybackslash\hyphenpenalty=10000\exhyphenpenalty=10000}X>{\hsize=0.591\hsize\raggedright\arraybackslash}X>{\hsize=1.104\hsize\raggedright\arraybackslash}X>{\hsize=1.076\hsize\raggedright\arraybackslash}X@{}}
\toprule
\textbf{Asset} & \textbf{Version} & \textbf{License} & \textbf{Citation (where applicable)} \\
\midrule
TransformerLens & 3.3.0 & MIT & \citealp{nanda2022transformerlens} \\
PyTorch & 2.12.0 & BSD 3-Clause & \citealp{paszke2019pytorch} \\
Transformers (Hugging Face) & 5.9.0 & Apache 2.0 & \citealp{wolf2020transformers} \\
safetensors & 0.7.0 & Apache 2.0 & NA \\
huggingface\_hub & 1.16.4 & Apache 2.0 & NA \\
NumPy & 2.4.6 & BSD 3-Clause & \citealp{harris2020numpy} \\
SciPy & 1.17.1 & BSD 3-Clause & \citealp{virtanen2020scipy} \\
pandas & 3.0.3 & BSD 3-Clause & \citealp{mckinney2010pandas} \\
scikit-learn & 1.8.0 & BSD 3-Clause & \citealp{pedregosa2011sklearn} \\
umap-learn & 0.5.12 & BSD 3-Clause & \citealp{mcinnes2018umap} \\
Matplotlib & 3.10.9 & Matplotlib License (PSF-based, BSD-compatible) & \citealp{hunter2007matplotlib} \\
Pillow & 12.2.0 & MIT-CMU & NA \\
adjustText & 1.4.0 & MIT & NA \\
tqdm & 4.67.3 & MPL-2.0 / MIT & NA \\
wordfreq (optional extra) & 3.1.1 & Apache-2.0 (code); CC BY-SA 4.0 (bundled data) & \citealp{speer2022wordfreq} \\
\bottomrule
\end{tabularx}
\end{minipage}
\end{center}

\begin{center}
\begin{minipage}{\linewidth}
\small
\captionof{table}{\textbf{Reproduced source materials.}}
\label{tab:E4}
\begin{tabularx}{\linewidth}{@{}>{\hsize=1.166\hsize\raggedright\arraybackslash\hyphenpenalty=10000\exhyphenpenalty=10000}X>{\hsize=0.804\hsize\raggedright\arraybackslash}X>{\hsize=1.030\hsize\raggedright\arraybackslash}X@{}}
\toprule
\textbf{Asset} & \textbf{License} & \textbf{Condition relevant to this release} \\
\midrule
\citet{sofroniew2026emotion} appendix materials (arXiv:2604.07729v1)—171-emotion vocabulary, 100-topic list, story- and neutral-dialogue generation prompts & No reuse license from either venue (arXiv non-exclusive distribution license; Transformer Circuits states no license)—authors retain copyright & Reproduced verbatim in Appendix A with attribution, as quotation for scholarly replication (fair use). The vocabulary and topic list are additionally non-original factual compilations; no license grant is relied upon. \\
\bottomrule
\end{tabularx}
\end{minipage}
\end{center}

\end{document}